\documentclass[10pt,twocolumn,letterpaper]{article}

\usepackage{bm}
\usepackage{iccv}
\usepackage{times}
\usepackage{epsfig}
\usepackage{graphicx}
\usepackage{amsmath}
\usepackage{amssymb}

\usepackage[pagebackref=true,breaklinks=true,letterpaper=true,colorlinks,bookmarks=false]{hyperref}
\usepackage[utf8]{inputenc} %
\usepackage[T1]{fontenc}    %
\usepackage{url}            %
\usepackage{booktabs}       %
\usepackage{amsfonts}       %
\usepackage{nicefrac}       %
\usepackage{microtype}      %
\usepackage{graphicx}
\usepackage{booktabs}
\usepackage[normalem]{ulem}
\usepackage{multirow}
\usepackage{amsmath}
\usepackage{rotating}
\usepackage{makecell}
\usepackage{stfloats}
\usepackage{wrapfig}
\usepackage{enumitem}
\usepackage{scalerel}
\usepackage{color}
\usepackage{gensymb}
\usepackage[dvipsnames]{xcolor}
\usepackage{scalerel}

\DeclareFontFamily{U}{mathb}{}
\DeclareFontShape{U}{mathb}{m}{n}{
  <-5.5> mathb5
  <5.5-6.5> mathb6
  <6.5-7.5> mathb7
  <7.5-8.5> mathb8
  <8.5-9.5> mathb9
  <9.5-11.5> mathb10
  <11.5-> mathb12
}{}
\DeclareSymbolFont{mathb}{U}{mathb}{m}{n}
\DeclareMathSymbol{\drsh}{3}{mathb}{"EB}

\newcommand{\round}[1]{\ensuremath{\lfloor#1\rceil}}

\newcommand\blfootnote[1]{%
  \begingroup
  \renewcommand\thefootnote{}\footnote{#1}%
  \addtocounter{footnote}{-1}%
  \endgroup
}

\definecolor{citecolor}{HTML}{0071bc}
\definecolor{gtred}{HTML}{FF3E30}
\definecolor{predblue}{HTML}{0776FF}
\hypersetup{
  breaklinks   = true,
  colorlinks   = true, %
  urlcolor     = RubineRed, %
  linkcolor    = RubineRed, %
  citecolor    = citecolor %
}

\iccvfinalcopy %

\def\iccvPaperID{****} %

\ificcvfinal\fi

\begin{document}

\title{Detector-Free Structure from Motion}

\author{
    Xingyi He$^{1}$,
    \quad Jiaming Sun$^{2}$ 
    \quad Yifan Wang$^{1}$ 
    \quad Sida Peng$^{1}$ 
    \\
    \quad Qixing Huang$^{3}$
    \quad Hujun Bao $^{1}$
    \quad Xiaowei Zhou$^{1\dagger}$
    \vspace{1em}
    \\
    $^1$Zhejiang University \quad 
    $^2$Image Derivative Inc. \quad
    $^3$The University of Texas at Austin \quad
}

\maketitle
\ificcvfinal\thispagestyle{empty}\fi

\blfootnote{The authors from Zhejiang University are affiliated with the State Key Lab of CAD\&CG. $^\dagger$Corresponding author: Xiaowei Zhou.}
\begin{abstract}
We propose a new structure-from-motion framework to recover accurate camera poses and point clouds from unordered images. 
Traditional SfM systems typically rely on the successful detection of repeatable keypoints across multiple views as the first step, which is difficult for texture-poor scenes, and poor keypoint detection may break down the whole SfM system.   
We propose a new detector-free SfM framework to draw benefits from the recent success of detector-free matchers to avoid the early determination of keypoints, while solving the multi-view inconsistency issue of detector-free matchers.
Specifically, our framework first reconstructs a coarse SfM model from quantized detector-free matches. Then, it refines the model by a novel iterative refinement pipeline, which iterates between an attention-based multi-view matching module to refine feature tracks and a geometry refinement module to improve the reconstruction accuracy. 
Experiments demonstrate that the proposed framework outperforms existing detector-based SfM systems on common benchmark datasets. We also collect a texture-poor SfM dataset to demonstrate the capability of our framework to reconstruct texture-poor scenes. 
Based on this framework, we take the \textbf{first place} in Image Matching Challenge 2023~\cite{imagematchingchallenge2023}.
Project page: \url{https://zju3dv.github.io/DetectorFreeSfM/}.
\end{abstract}
\section{Introduction}

Structure-from-Motion~(SfM) is a fundamental task in computer vision, which aims to recover camera poses, intrinsic parameters, and point clouds from multi-view images of a scene.
The estimated camera poses and optional point clouds benefit downstream tasks, such as visual localization, multi-view stereo, and novel view synthesis.

\begin{figure}[tp]
    \centering
    \includegraphics[width=0.9\linewidth]{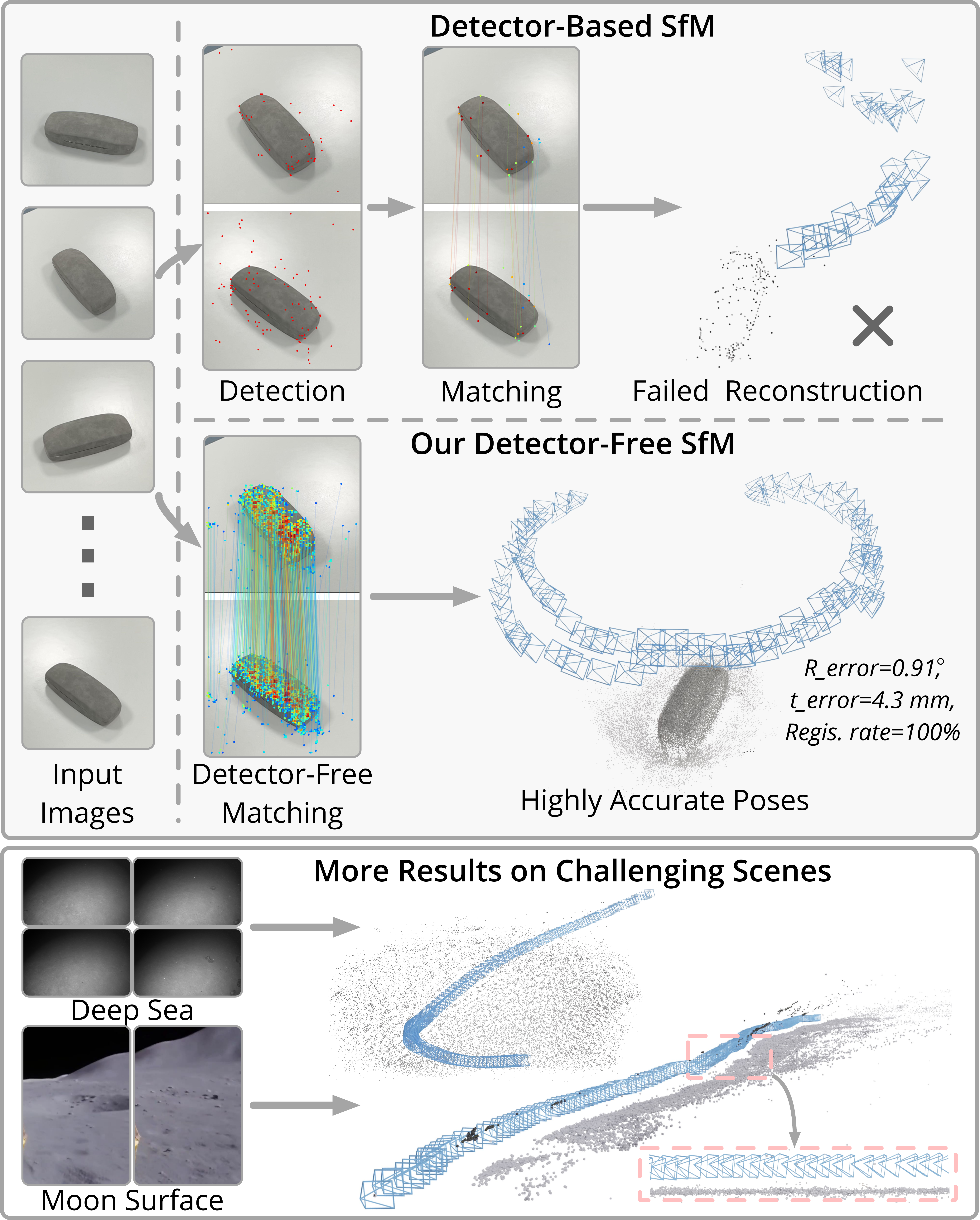}
    \caption{\textbf{Comparison between traditional detector-based SfM and the proposed detector-free SfM.}
    For the texture-poor scene, detector-based SfM fails due to the poor repeatability of detected keypoints at the beginning, while our detector-free SfM framework leverages detector-free matching and achieves complete reconstruction with highly accurate camera poses.
    Our framework is applicable to real-world challenging scenes such as the deep sea and the moon surface.
    }
    \vspace{-0.55 cm}
    \label{fig:overview}
\end{figure}

SfM has been studied for decades, with many well-established methods~\cite{Agarwal2009BuildingRI,Crandall2011DiscretecontinuousOF,Wilson2014RobustGT,Cui2015GlobalSB}, open-source systems such as Bundler~\cite{Snavely2006PhotoTE} and COLMAP~\cite{schonbergerStructurefromMotionRevisited2016}, and commercial software~\cite{realtycapture,Metashape} that are accurate and scalable to large-scale scenes.
As a routine, they require to detect and match sparse feature points across multiple views~\cite{LoweDavid2004DistinctiveIF,DeTone2017SuperPointSI,sarlin20superglue} at the beginning of the pipeline to build multi-view point-to-point correspondences.
This requirement could not be fulfilled in many cases. For example, in texture-poor regions, it is hard to robustly detect repeatable keypoints across multiple views for matching.
The poor feature detection and matching become the bottleneck of the whole SfM pipeline, which leads to missing image registration or even failed reconstruction of the entire model.
Fig.~\ref{fig:overview} presents an example.

\begin{figure}[tp]
    \centering
    \includegraphics[width=0.7\linewidth]{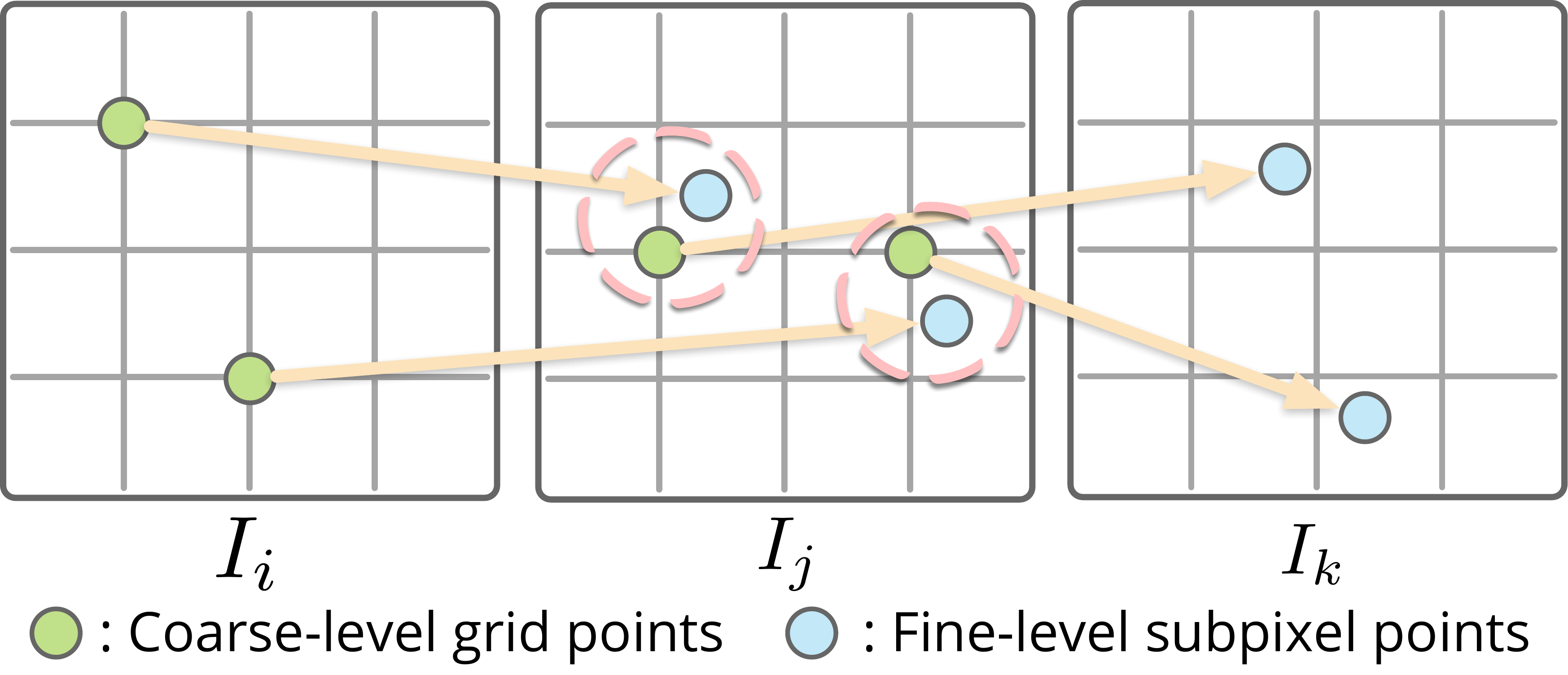}
    \caption{\textbf{Multi-view Inconsistency Issue of Detector-Free Matching.} The resulting feature locations of $I_j$ are varied when $I_j$ is matched to $I_i$ and $I_k$, yielding fragmentary feature tracks.
    }
    \vspace{-0.55 cm}
    \label{fig:nonrepeatable}
\end{figure}

Recently, detector-free matchers~\cite{Sun2021LoFTRDL,Chen2022ASpanFormerDI,wang2022matchformer} achieve state-of-the-art performance on the image-matching task.
They have shown a strong capability for matching low-textured regions with the help of the detector-free design and the attention mechanism~\cite{Vaswani2017AttentionIA}.
They often use a coarse-to-fine matching strategy for efficiency.
The dense matching on a coarse grid is first performed between downsampled feature maps of two images.
Then, the feature locations of coarse matches on one image are fixed, while their subpixel correspondences are searched on the other image with fine-level feature maps.
Therefore, the resulting feature locations in an image depend on the other image, as shown in Fig.~\ref{fig:nonrepeatable}. 
This pair-dependent nature leads to fragmentary feature tracks when running pair-wise matching over multiple views, which makes detector-free matchers not directly applicable to existing SfM systems.

In this paper, we propose a new SfM framework that is able to leverage the recent success of detector-free matching and recover highly-accurate camera poses even for texture-poor scenes.
An overview of our pipeline is depicted in Fig.~\ref{fig:allpipeline}.
To solve the inconsistency issue of detector-free matching, our SfM framework reconstructs the scene in a coarse-to-fine manner, which first builds a coarse SfM model with the quantized matches, and then iteratively refines the model towards higher accuracy.

Specifically, our framework first matches image pairs with a detector-free feature matcher, e.g., LoFTR~\cite{Sun2021LoFTRDL}. 
Then, in the coarse reconstruction phase, we quantize the feature locations by rounding them into a coarse grid to improve consistency and reconstruct a coarse SfM model.
This coarse model provides initial camera poses and scene structures for the later refinement phase.
Next, we propose an iterative refinement pipeline that alternates between a feature track refinement phase and a geometry refinement phase to improve pose and point cloud accuracy.
The feature track refinement module is built on a novel transformer-based multi-view matching network, which enhances the discrimitiveness of extracted features by encoding positional and multi-view context with self- and cross-attention mechanisms.
Based on refined feature tracks, the geometry refinement module uses bundle adjustment and track topology adjustment to improve the accuracy of camera poses and point clouds.

Experiments on the public ETH3D dataset~\cite{Schps2017AMS} and Image Matching Challenge~(IMC)~\cite{jin2021image} dataset demonstrate that our detector-free SfM framework outperforms state-of-the-art detector-based SfM systems with respect to various metrics.
To further evaluate and demonstrate the capability of our SfM framework on challenging scenes, we also collect a texture-poor SfM dataset which is composed of 17 scenes with 1020 image bags.
Thanks to the detector-free design and the iterative refinement pipeline, our framework can recover accurate camera poses with high registration rates even for challenging texture-poor scenes.
Fig.~\ref{fig:overview} presents some examples.

\paragraph{Contributions:}
\vspace{-.4em}
\begin{itemize}[leftmargin=*]
\setlength\itemsep{-.3em}

\item A new detector-free SfM framework built upon detector-free matchers to handle texture-poor scenes.
\item An iterative refinement pipeline with a transformer-based multi-view matching network to efficiently refine both feature tracks and reconstruction results.
\item A new texture-poor SfM dataset with ground-truth pose annotations.
\end{itemize}
\section{Related Work}

\paragraph{Structure-from-Motion.}
Feature correspondence-based SfM methods have long been investigated~\cite{Mohr1993Relative3R,Beardsley19963DMA,Fitzgibbon1998AutomaticCR,Pollefeys2004VisualMW}.
Many previous works focus on improving the efficiency and robustness of large-scale scene reconstruction~\cite{Agarwal2009BuildingRI,Agarwal2010BundleAI,Crandall2011DiscretecontinuousOF,Wilson2014RobustGT,Cui2015GlobalSB,schonbergerStructurefromMotionRevisited2016}.
Some methods try to disambiguate matches when applied to scenes with highly repetitive or symmetric structures~\cite{Roberts2011StructureFM, Wilson2013NetworkPF}.
As discussed in the introduction, these methods require feature detection and matching at the beginning of the pipeline.
In challenging scenes, especially in texture-poor regions, poor keypoint detection will affect the overall SfM pipeline.

More recent end-to-end SfM methods propose to directly regress poses~\cite{Vijayanarasimhan2017SfMNetLO,Zhou2017UnsupervisedLO,Parameshwara2022DiffPoseNetDD,zhang2022relpose} or solve poses using differential bundle adjustment~(BA)~\cite{Tang2019BANetDB,Gu2021DRODR}.
These methods avoid explicit feature matching and thus don't suffer from poor feature matching.
However, they have limited scalability and generalizability on real-world settings. 
With the success of recent neural scene representations, some methods~\cite{Lin2021BARFBN,Jeong2021SelfCalibratingNR} try to optimize poses with differentiable rendering.
However, they often rely on using previous correspondence-based methods, e.g., COLMAP~\cite{schonbergerStructurefromMotionRevisited2016}, to provide initial poses, as joint pose and scene optimization from scratch is difficult to converge and prone to local minima, c.f.~\cite{Lin2021BARFBN,Meng2021GNeRFGN}.

Different from these previous methods, our detector-free SfM framework eliminates the requirement of sparse feature detection at the beginning of the pipeline, which is more robust in challenging scenarios such as low-textured regions and repetitive patterns.
Moreover, our framework is scalable to large-scale scenes and can handle in-the-wild data with extreme view-point and illumination changes.
~\cite{Widya2018StructureFM,he2022oneposeplusplus} are relevant to our framework which also eliminates feature detection by performing coarse grid-level matching first and then refining 2D points for sub-pixel accuracy.
Different from their refinement that is single- or two-view based, our framework is capable of leveraging multi-view information to refine a feature track.

\paragraph{Feature Matching.} 
Feature Matching is often a prerequisite for SfM and SLAM.
A typical feature matching pipeline~\cite{LoweDavid2004DistinctiveIF,Rublee2011ORBAE,DeTone2017SuperPointSI,Dusmanu2019CVPR,r2d2} is to detect and describe keypoints on each image, and then match them by nearest neighbor search or learning-based matchers~\cite{sarlin20superglue,Chen2021LearningTM}.
The merit of these methods is the high matching efficiency based on the sparse points.
However, for challenging scenarios, especially low-textured regions, poor feature detection at the beginning is the bottleneck and affects the overall SfM system.

In recent years, many methods directly match image pairs in a dense~\cite{Truong2021LearningAD} or semi-dense manner~\cite{Rocco2018NeighbourhoodCN,li20dualrc,Sun2021LoFTRDL,tang2022quadtree,Chen2022ASpanFormerDI,wang2022matchformer}, avoiding feature detection.
With the help of Transformer~\cite{Vaswani2017AttentionIA}, some semi-dense matching methods~\cite{Sun2021LoFTRDL,Chen2022ASpanFormerDI,wang2022matchformer} achieve higher accuracy compared with detector-based baselines and show strong capabilities in building correspondences on low-textured regions.
However, due to their inconsistency problem when matching multiple views (shown in Fig.~\ref{fig:nonrepeatable}), it is hard to directly apply them to the current SfM systems, as discussed in the introduction.
While rounding~\cite{Chen2022ASpanFormerDI} or merging strategies~\cite{Shen2022SemiDenseFM} could be used to produce long feature tracks for SfM, these strategies sacrifice the matching accuracy, which will significantly reduce the accuracy of the reconstructed SfM models.
Unlike them, our detector-free SfM framework with a coarse-to-fine manner can recover highly accurate poses and point clouds.

\paragraph{Multi-View Refinement.}
Accurate multi-view correspondences are crucial for recovering accurate point clouds and camera poses in SfM.
The technical challenge is that per-view detection of feature points cannot guarantee their geometric consistency among multiple views.
To solve this problem, some previous methods perform multi-view refinement with flow~\cite{Dusmanu2020MultiViewOO} or dense features~\cite{Lindenberger2021PixelPerfectSW}, which bring significant accuracy improvement for SfM.
PatchFlow~\cite{Dusmanu2020MultiViewOO} first estimates the dense flow field within the local patch of each tentative pair and then refines multi-view 2D locations by minimizing the energy function based on the estimated flow.
PixSfM~\cite{Lindenberger2021PixelPerfectSW} performs feature-metric keypoint adjustment and bundle adjustment to refine 2D feature locations before SfM and the entire scene after SfM, respectively.
Our detector-free SfM framework may adopt these two methods to refine the quantized matches and SfM models.
However, PatchFlow suffers from high computation due to pair-wise flow estimations.
PixSfM needs to preserve feature patches or cost maps of all 2D observations in the memory for the feature-metric BA.
Given that detector-free matchers produce significantly more correspondences than sparse matchers, the memory footprint of adapting PixSfM to our detector-free SfM pipeline is inevitably large, especially on large-scale scenes.
Different from them, we devise a transformer-based multi-view refinement matching module, which can efficiently and accurately refine a feature track with a single forward pass.
Moreover, thanks to the design of our refinement phase that separately refines feature tracks and performs geometry refinement, the geometric BA can be leveraged for efficiency both in terms of speed and memory.
Experimental comparisons are provided in Sec.~\ref{subsec scalability}.

\section{Method}
\begin{figure*}[tp]
    \centering
    \includegraphics[width=1.0\linewidth]{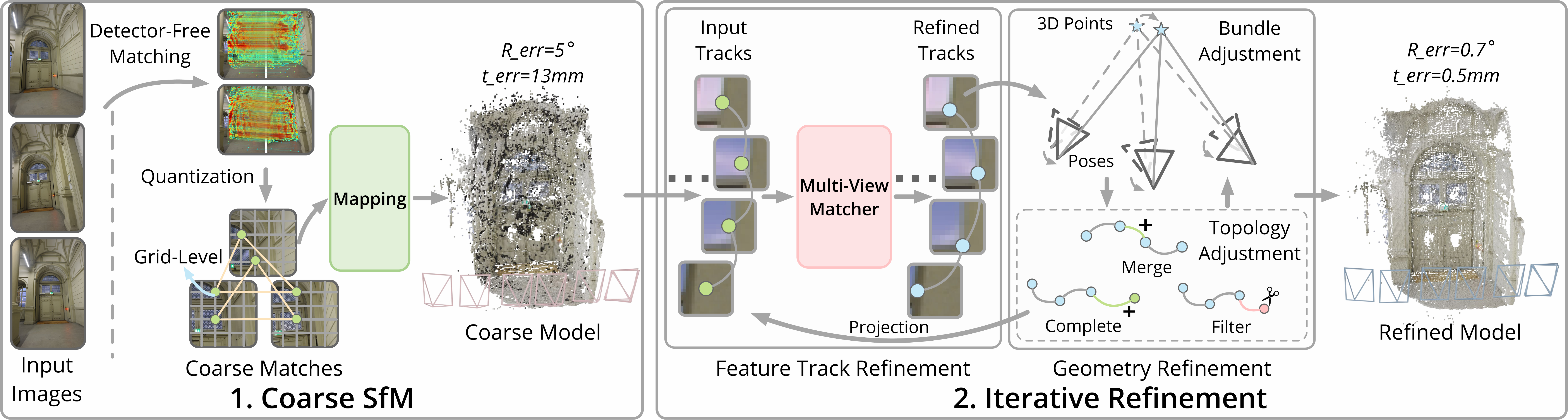}
    \caption{\textbf{Pipeline Overview.} 
    Beginning with a collection of unordered images, the \textbf{Coarse SfM} stage generates an initial SfM model based on multi-view matches from a detector-free matcher.
    Then, the \textbf{Iterative Refinement} stage improves the accuracy of the SfM model by alternating between the feature track refinement module and the geometry refinement module.
    }
    \label{fig:allpipeline}
    \vspace{-0.35 cm}
\end{figure*}

An overview of our detector-free SfM framework is shown in Fig.~\ref{fig:allpipeline}.
Given a set of unordered images $\{\*I_i\}$, our objective is to recover camera poses $\{\boldsymbol{\xi}_i \in \mathbb{SE}(3) \}$, intrinsic parameters $\{\*C_i\}$ and a scene point cloud $\{\*P_j\}$. 
The recovered camera poses are in a global coordinate system.
To achieve this goal, we propose a two-stage pipeline, in which we first establish correspondences between image pairs with a detector-free matcher and reconstruct an initial coarse SfM model~(Sec.~\ref{subsec:coarse}). 
Then, we perform iterative refinement to improve the accuracy of poses and point clouds (Sec.~\ref{subsec:fine}).

\subsection{Detector-Free Matching and Coarse SfM}
\label{subsec:coarse}

For a set of unordered images, our framework directly performs detector-free semi-dense feature matching between image pairs instead of first detecting sparse keypoints as in traditional SfM pipeline~\cite{schonbergerStructurefromMotionRevisited2016}.
Eliminating the keypoint detection phase can help avoid poor detection affecting the overall SfM system and benefit the reconstruction of challenging texture-poor scenes.

\paragraph{Match Quantization.}
Directly adapting the correspondences of semi-dense matchers for SfM is not straightforward, due to the inconsistent multi-view matches as depicted in Fig.~\ref{fig:nonrepeatable} and discussed in the introduction.
Our idea is to strive for match consistency by sacrificing accuracy in the coarse SfM phase.
Concretely, we quantize the 2D locations of matches into a grid:~$\round{\*x/r}*r$, where $\round{\cdot}$ is the rounding operator and $r$ is the grid cell size.
This quantization step forces multiple subpixel matches that are close to each other to merge into a single grid node, which improves consistency.
Note that the coarse-level correspondences output by some detector-free~\cite{Sun2021LoFTRDL,wang2022matchformer,Chen2022ASpanFormerDI} matchers are typically at $\nicefrac{1}{8}$ image resolution, which can directly be used as quantized matches.
The ablation analysis of $r$ is given in Sec.~\ref{sec:abl}.

After the match quantization, we utilize these coarse matches for incremental mapping~\cite{schonbergerStructurefromMotionRevisited2016} to obtain a coarse SfM model. 
The accuracy of recovered camera poses and point clouds are limited due to the match quantization, which serves as the initialization of our refinement framework introduced in the next section.

\subsection{Iterative SfM Refinement}
\label{subsec:fine}
We proceed to refine the initial SfM model to obtain improved camera poses and point clouds.
To this end, we propose an iterative refinement pipeline.
Within each iteration, we first enhance the accuracy of feature tracks with a multi-view matching module.
These refined feature tracks are then fed into a geometry refinement phase which optimizes camera poses and point clouds jointly.
The refinement process can be performed multiple times for higher accuracy. An overview is shown in Fig.~\ref{fig:allpipeline}.

\subsubsection{Feature Track Refinement}
\begin{figure}[tp]
    \centering
    \includegraphics[width=\linewidth]{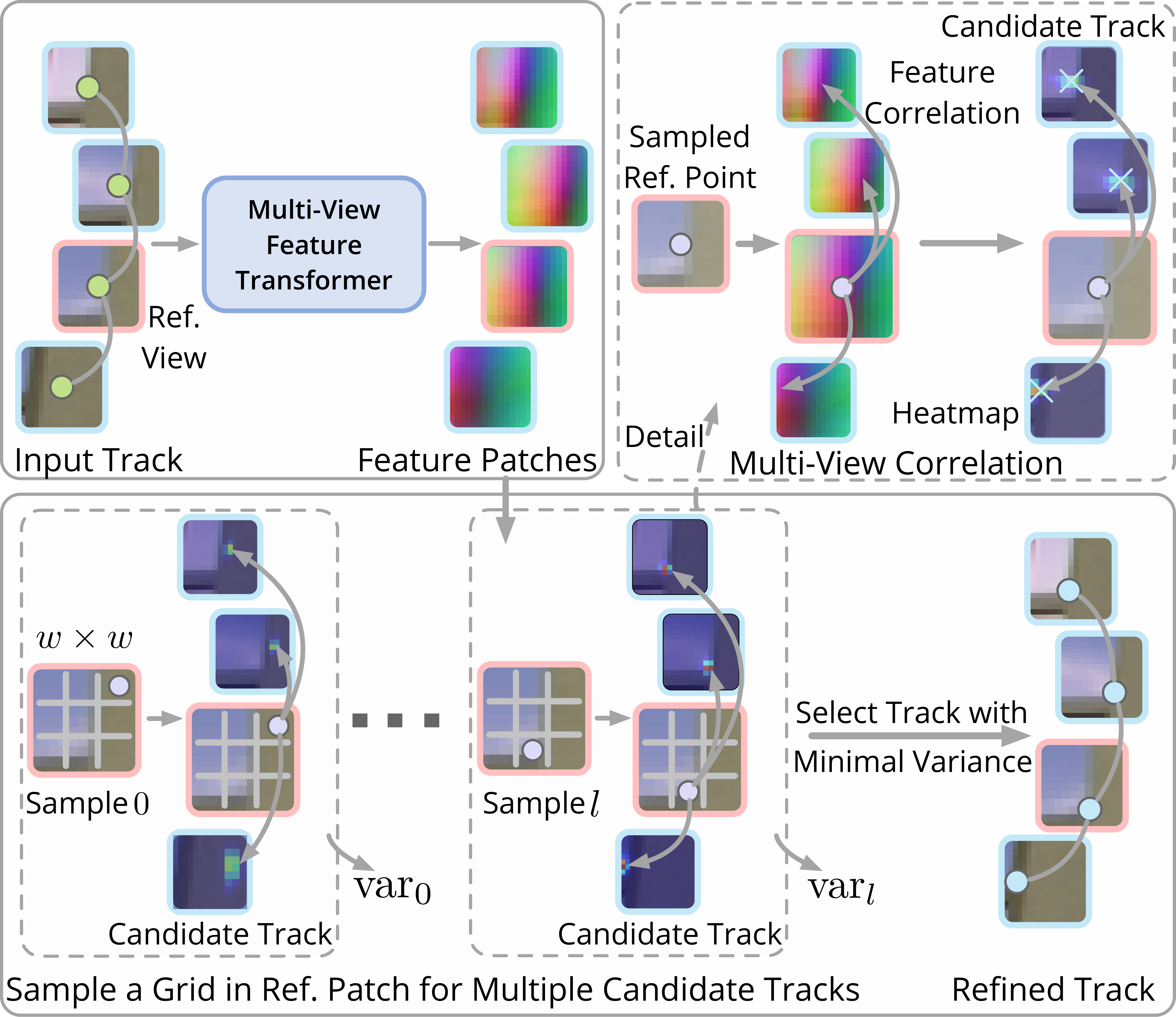}
    \caption{\textbf{Multi-View Matching Module.}
    Given an input feature track with a selected reference view~(\protect\scalerel*{\includegraphics{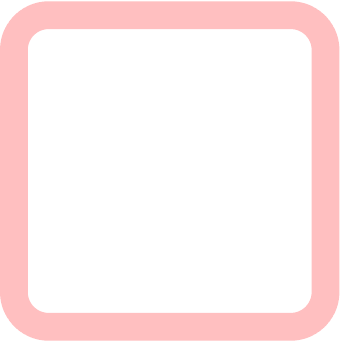}}{B}), the local patches centered at the keypoints are fed into a multi-view feature transformer to extract feature patches.
    A $w \times w$ grid of reference locations is sampled in the reference view. For each reference location~(\protect\scalerel*{\includegraphics{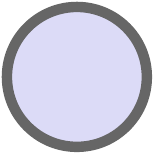}}{B}), its feature is correlated with the feature patches of query views~(\protect\scalerel*{\includegraphics{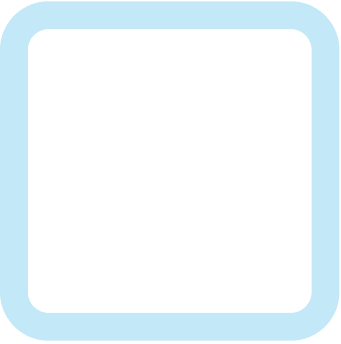}}{B}) to obtain heatmaps that indicate the expected keypoint locations and their variances in the query views, yielding a candidate feature track.
    This process is repeated for all reference locations.   
    Finally, the candidate track with the smallest variance is selected as the refined track.
    }
    \vspace{-0.35 cm}
    \label{fig:trackrefine}
\end{figure}

A feature track $\mathcal{T}_j = \{\mathbf{x}_k \in \mathbb{R}^2 | k=1:N_j\}$ is a set of 2D keypoint locations in multi-view images corresponding to a 3D scene point $\*P_j$.
We devise a multi-view matching module to efficiently refine feature tracks $\{\mathcal{T}_j\}$ for high accuracy, which is illustrated in Fig.~\ref{fig:trackrefine}.
The basic idea is to locally adjust the keypoint locations in all views so that the correlation among their features is maximized. 

As exhaustively correlating all pairs of views is computationally intractable, we select a reference view, extract the feature at the keypoint in the reference view, and correlate it with the local feature maps with a size of $p\times p$ around the keypoints in other views (called query views), yielding a set of $p\times p$ heatmaps that can be viewed as distributions of the keypoint locations. In each query view, we compute the expectation and variance over each heatmap as the refined keypoint location and its uncertainty, respectively. 
This process gives us a candidate feature track with refined keypoint locations in all query views as well as the uncertainty of this candidate track, i.e., the sum of variance over all the heatmaps. 
To also refine the keypoint location in the reference view, we sample a $w \times w$ grid of reference locations around the original keypoint in the reference view. Then, we repeat the above feature correlation procedure to produce a candidate feature track for each sampled reference location. 
Finally, the candidate track with the smallest uncertainty is selected as the refined feature track $\mathcal{T}_j^*$.

\paragraph{Reference View Selection.}
For each feature track, our criteria to select the reference view is to minimize the keypoint scale differences between the reference view and query views to improve the matchability.
Specifically, we compute the depth values of keypoints based on the currently recovered poses and point clouds, which indicate the scale information. 
Then, the view with a medium scale across the track is selected as the reference view whereas the rest views are query views.
More details about scale estimation can be found in the supplementary material.

\begin{figure}[tp]
    \centering
    \includegraphics[width=\linewidth]{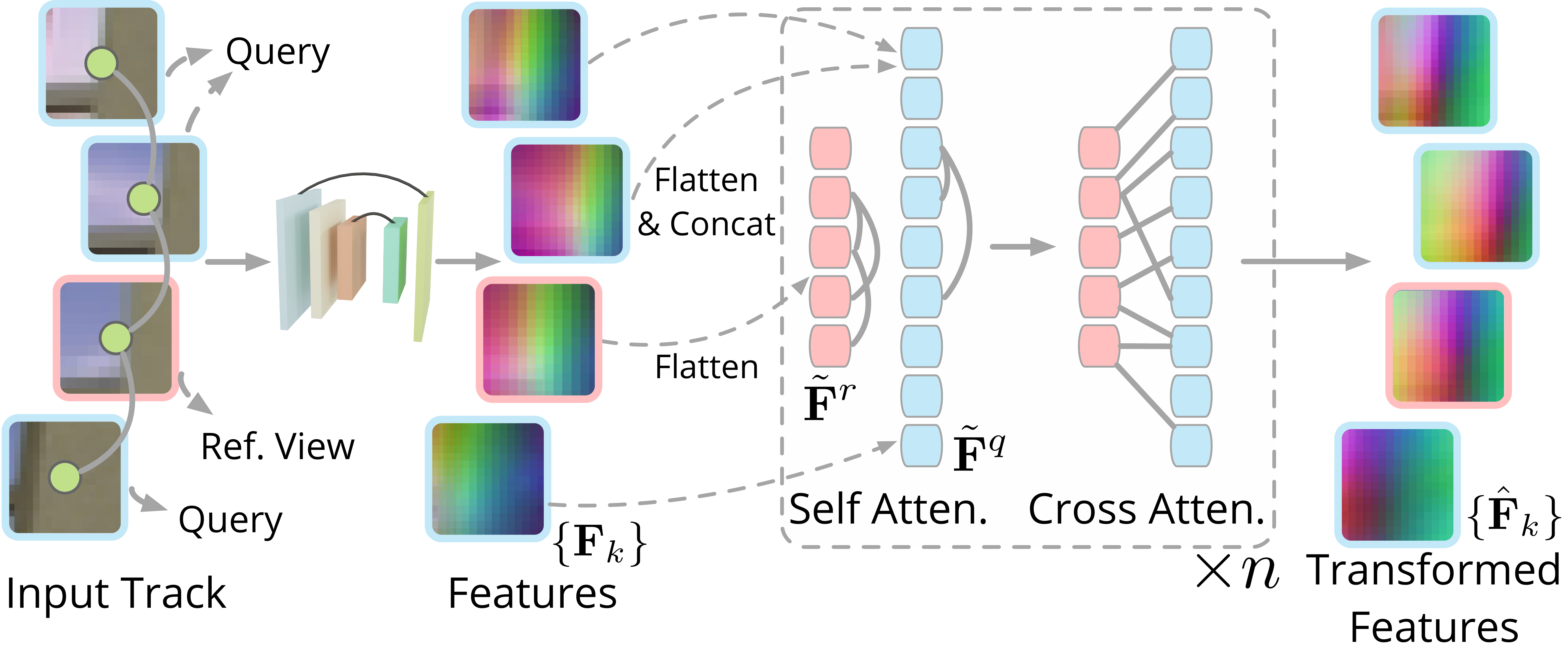}
    \caption{\textbf{Multi-View Feature Transformer.}
    The local patches centered at the keypoints of an input feature track are fed into a CNN to extract features and then flattened and concatenated to perform multiple self- and cross-attentions.
    }
    \vspace{-0.35 cm}
    \label{fig:model_detail}
\end{figure}

\paragraph{Multi-View Feature Transformer.}
The multi-view matching needs to extract local feature patches centered at 2D keypoints of each $\mathcal{T}_j$.
Instead of using a CNN, we design a multi-view feature transformer to enhance the discrimitiveness of extracted features by encoding multi-view context with attention mechanisms. 
As shown in Fig. \ref{fig:model_detail}, we feed the $p \times p$ image patches centered at each keypoint into a CNN backbone to obtain a set of feature patches $\{\mathbf{F}_k \in \mathbb{R}^{p \times p \times c} \}$, where $c$ is the number of channels. Then, $\{\mathbf{F}_k\}$ are flattened to $\{\tilde{\mathbf{F}}_k \in \mathbb{R}^{m \times c}\}$, where $m = p \times p $.
The flattened features of query views are concatenated into a single query feature $\tilde{\mathbf{F}}^q$ along the first dimension.
Then, we perform self- and cross-attention by $n$ times between flattened reference feature $\tilde{\mathbf{F}}^r$ and query feature $\tilde{\mathbf{F}}^q$ to obtain the transformed multi-view features $\{\hat{\mathbf{F}}_k\}$, which are used for feature correlation to refine the feature track.

\paragraph{Training.}
Besides the detector-free matcher, the only learned module in our framework is the multi-view feature transformer. It is trained on MegaDepth~\cite{Li2018MegaDepthLS} by minimizing the average $\ell_2$ loss on keypoint locations between the refined tracks and the ground-truth tracks.
We construct training data by sampling image bags on each scene with a maximum of six images in each bag.
Image bags are sampled by the co-visibility extracted from the provided scene SfM model.
Then, the ground-truth feature tracks in each bag are built by randomly selecting a reference image and projecting its grid points to other views by depth maps.
The 2D locations of tracks in the query views are perturbed randomly by a maximum of seven pixels to generate the coarse feature tracks, which are the input of our multi-view matching module.
More details are provided in the supplementary material.

\subsubsection{Geometry Refinement}

Based on the previous refined feature tracks $\{\mathcal{T}_j^* \}$, our geometry refinement pipeline iteratively refines the poses, intrinsics, point clouds, as well as topology of feature tracks.
Track topology means the graph structure of a set of connected 2D keypoints.

Unlike PixSfM~\cite{Lindenberger2021PixelPerfectSW} that needs to preserve feature patches or cost maps of all 2D observations in memory to perform feature-metric BA, we can directly perform the efficient geometric BA~\cite{Triggs1999BundleA} to optimize poses and point clouds based on the refined feature tracks.
Formally, we minimize the reprojection error to optimize intrinsic parameters~$\{\*C_i\}$, poses~$\{\boldsymbol{\xi}_i\}$, and 3D points~$\{\*P_j\}$:
\begin{equation*}
    \resizebox{0.8\linewidth}{!}{
    $
    E = \sum_j \sum_{\*x_k^* \in \mathcal{T}^*_j} \rho \bigl( \left\lVert \pi \left(\boldsymbol{\xi}_i \cdot \*P_j, \*C_i\right) - \*x^*_k \right\rVert_2^2 \bigr) \enspace,
    $
    }
    \label{eq:gba}
\end{equation*}
where $\pi(\cdot)$ project points in the camera coordinate to image plane by $\*C_i$, $\rho(\cdot)$ is a robust loss function~\cite{hampel1986robust}.

After BA, we perform the feature track topology adjustment~(TA) based on the refined model, which benefits further BA and multi-view matching.
Since the overall scene is more accurate after the multi-view refinement and BA, we adjust the topology of feature tracks by adding 2D keypoints that previously failed to be registered into feature tracks and merging the tracks that can meet the reprojection criteria at this time, following~\cite{Wu2013TowardsLI,schonbergerStructurefromMotionRevisited2016}.
The outlier filtering~\cite{Snavely2006PhotoTE, Wu2013TowardsLI,schonbergerStructurefromMotionRevisited2016} is also performed to further reject points that cannot meet the maximum reprojection threshold $\epsilon$ after the refinement.

We alternate BA and TA multiple times to obtain the refined poses and point clouds.
Then, we project the refined point clouds to images with the current poses to update their 2D locations, which will serve as the initialization of the multi-view matching in the next refinement iteration.

\subsection{Texture-Poor SfM Dataset}
\label{ourdataset}

We collect an SfM dataset composed of $17$ object-centric texture-poor scenes with accurate ground-truth poses.
In our dataset, low-textured objects are placed on a texture-less plane.
For each object, we record a video sequence of around 30 seconds surrounding the object. The per-frame ground-truth poses are estimated by ARKit~\cite{arkit} and BA post-processing, with the assistance of textured markers, which are cropped out in test images.
To impose larger viewpoint changes, we sample $60$ subset image bags for each scene, similar to the IMC dataset~\cite{jin2021image}.
Example images are shown in Fig.~\ref{fig:reconresults} and more details are in the supplementary material.

\section{Experiments}

\subsection{Baselines and Datasets}

\paragraph{Baselines.}
We compare our method with a few baseline methods in two categories: 1) Detector-based SfM pipeline~\cite{schonbergerStructurefromMotionRevisited2016} with different features, including SIFT~\cite{LoweDavid2004DistinctiveIF}, D2-Net~\cite{Dusmanu2019CVPR}, R2D2~\cite{r2d2} and SuperPoint~(SP)~\cite{DeTone2017SuperPointSI}, and matchers, including Nearest Neighbor ~(NN) and SuperGlue~(SG)~\cite{sarlin20superglue}.
All these detector-based baselines are coupled with PixSfM~\cite{Lindenberger2021PixelPerfectSW}, which is the state-of-the-art SfM refinement method.
2) Detector-free SfM baseline LoFTR~\cite{Sun2021LoFTRDL} + PixSfM~\cite{Lindenberger2021PixelPerfectSW}, where PixSfM is fed with LoFTR's matches, which are quantized by the same strategy as in our pipeline.
\begin{table*}[t]
    \centering
    \resizebox{0.9\textwidth}{!}{
    \setlength\tabcolsep{6pt} %
    \begin{tabular}{ccccccccccc} 
    \toprule
    \multirow{2}{*}{Type} & \multirow{2}{*}{Method}         & \multicolumn{3}{c}{ETH3D Dataset}             & \multicolumn{3}{c}{IMC Dataset} & \multicolumn{3}{c}{Texture-Poor SfM Dataset}   \\ 
    \cmidrule(lr){3-5}
    \cmidrule(lr){6-8}
    \cmidrule(lr){9-11}
        &              & AUC@1$\degree$       & AUC@3$\degree$       & AUC@5$\degree$  & AUC@3$\degree$       & AUC@5$\degree$       & AUC@10$\degree$   & AUC@3$\degree$ & AUC@5$\degree$ & AUC@10$\degree$ \\ 
    \midrule
    \multirow{5}{*}{Detector-Based} &   COLMAP~(SIFT+NN) & 26.71 & 38.86 & 42.14 & 23.58 & 32.66 & 44.79 & 2.87 & 3.85 & 4.95 \\
    &   SIFT + NN + PixSfM & 26.94 & 39.01 & 42.19 & 25.54 & 34.80 & 46.73 & 3.13 & 4.08 & 5.09\\
    &   D2Net + NN + PixSfM & 34.50 & 49.77 & 53.58 & 8.91 & 12.26 & 16.79 & 1.54 & 2.63 & 4.54\\
    &   R2D2 + NN + PixSfM & 43.58 & 62.09 & 66.89 & 31.41 & 41.80 & 54.65 & 3.79 & 5.51 & 7.84\\
    &   SP + SG + PixSfM & 50.82 & 68.52 & 72.86 & 45.19  & 57.22 & 70.47 & 14.00 & 19.23 & 24.55\\
    \hline
    \multirow{4}{*}{Detector-Free} & LoFTR + PixSfM & 54.35 & 73.97 & 78.86 & 44.06 & 56.16 & 69.61 & 20.66 & 30.49 & 42.01\\
    & Ours~(LoFTR) & \textbf{59.12} & \textbf{75.59} & \textbf{79.53} & \underline{46.55}  & \underline{58.74} & \underline{72.19} & \underline{26.07} & 35.77 & 45.43 \\
    & Ours~(AspanTrans.) & \underline{57.23} & \underline{73.71} & \underline{77.70} & \textbf{46.79} & \textbf{59.01} & \textbf{72.50} & 25.78 & \underline{35.69} & \underline{45.64} \\
    & Ours~(MatchFormer) & 56.70 & 73.00 & 76.84 & 45.83 & 57.88 & 71.22 & \textbf{26.90} & \textbf{37.57} & \textbf{48.55}\\
    \bottomrule
    \end{tabular}
    }
    \vspace{0.15cm}
    \caption{\textbf{Results of Multi-View Camera Pose Estimation.}
    Our framework is compared with detector-based and detector-free baselines on multiple datasets by the AUC of pose error at different thresholds. 
    \textbf{Bold} and \underline{underline} indicate the best and second-best results.
    }
    \label{tab:exp camerapose}
    \vspace{-0.35 cm}
    \end{table*}
\begin{figure*}[btp]
	\centering
	\includegraphics[width=0.95\linewidth]{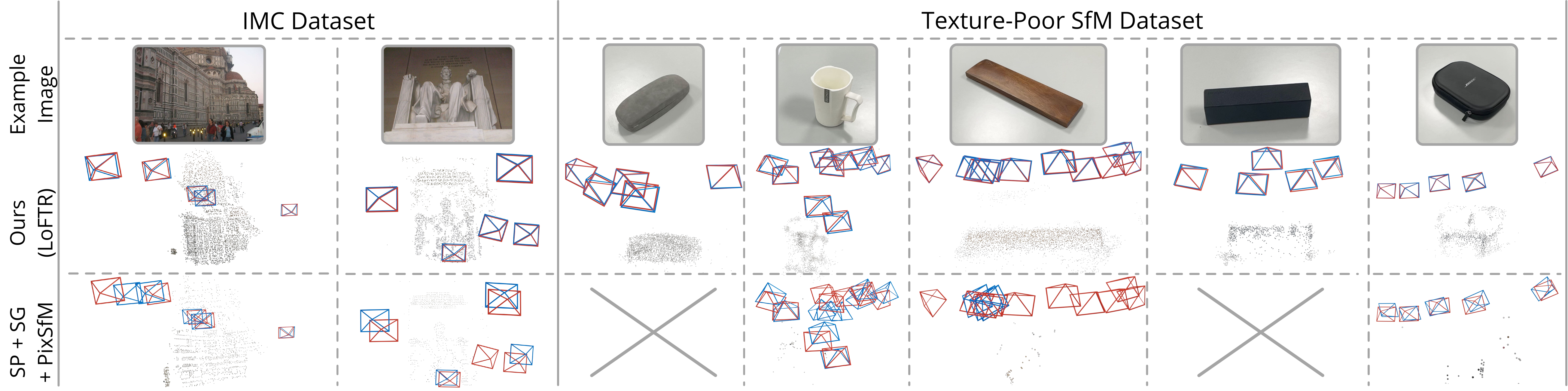}
	\caption{{\textbf{Qualitative Results.}}
	Our method with detector-free matcher LoFTR~\cite{Sun2021LoFTRDL} is qualitatively compared with the detector-based baseline \textit{SP + SG + PixSfM} on multiple scenes. The {red} cameras~(\protect\scalerel*{\includegraphics{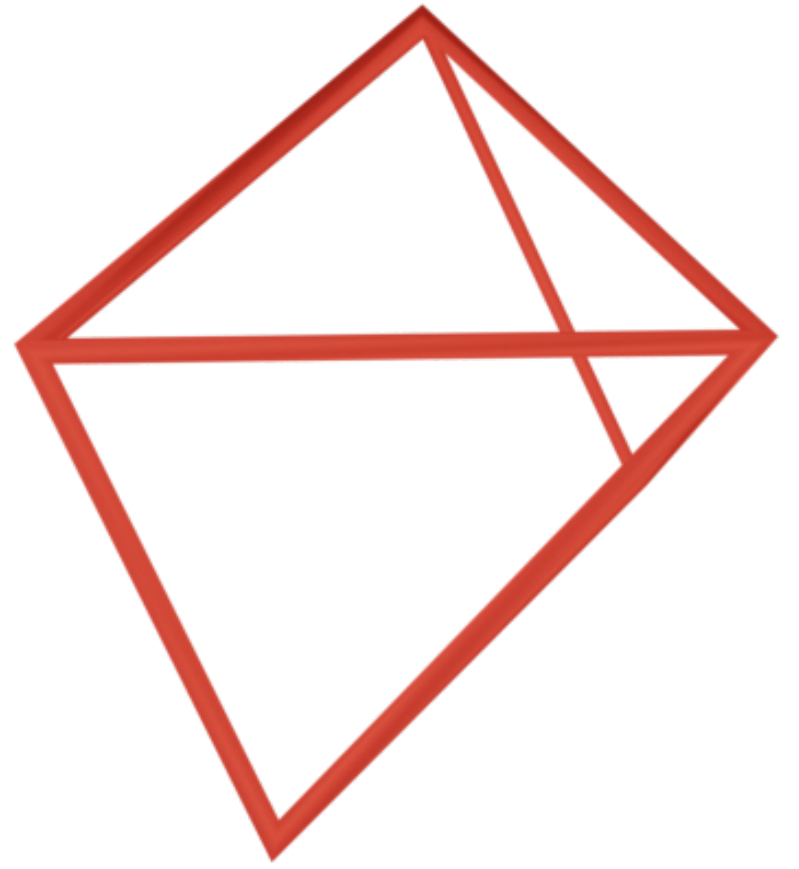}}{B}) are ground-truth poses while the {blue} cameras~(\protect\scalerel*{\includegraphics{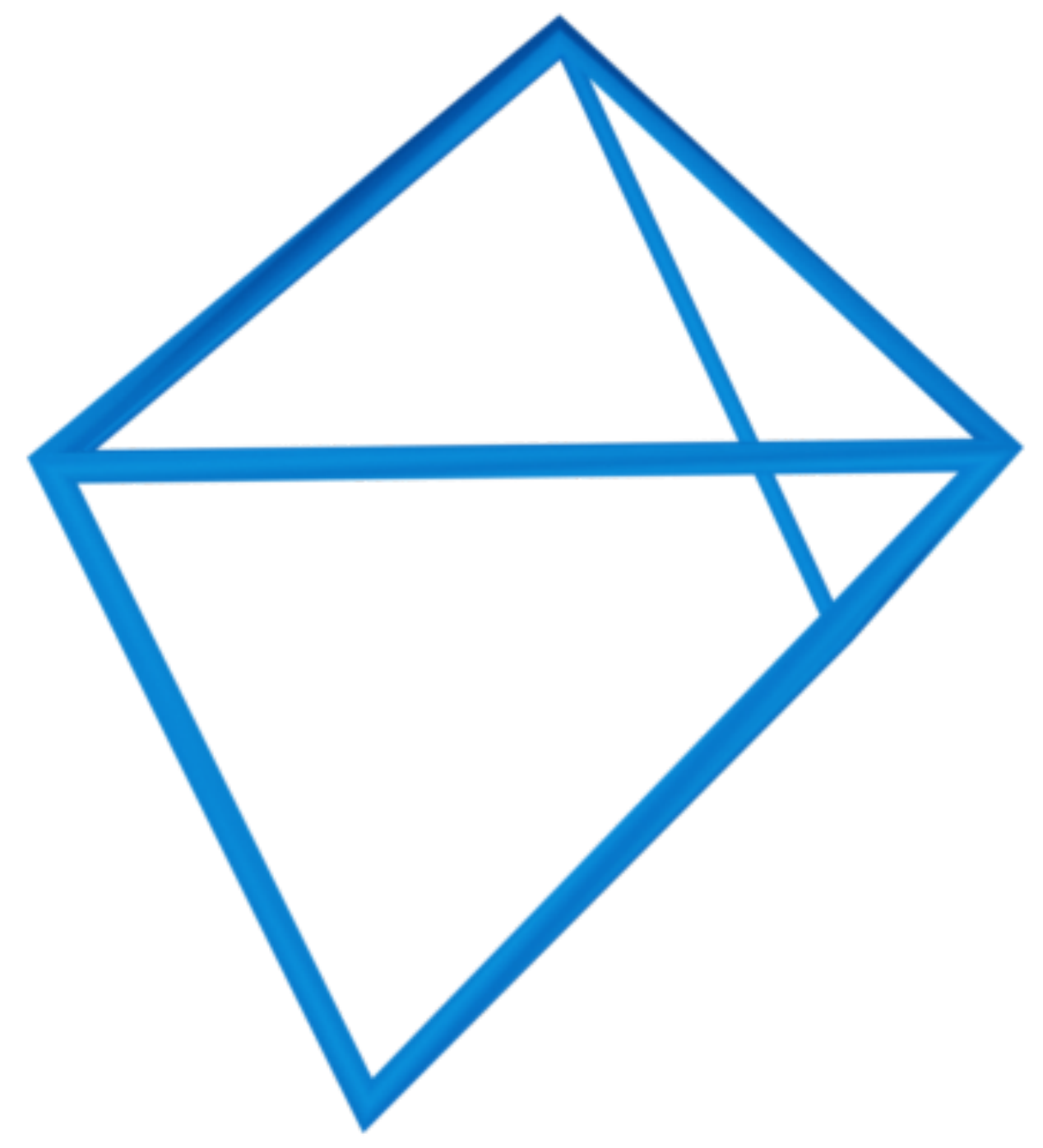}}{B}) are recovered poses.
	}
	\label{fig:reconresults}
	  \vspace{-0.3 cm}
\end{figure*}

\paragraph{Datasets.} Datasets used for evaluation include the Image Matching Challenge~(IMC) 2021 dataset~\cite{jin2021image}, the ETH3D dataset~\cite{Schps2017AMS}, and the proposed Texture-Poor SfM dataset. 
These datasets cover multiple types of scenes with different challenges.
The IMC Phototourism dataset contains large-scale outdoor scenes. All eight test scenes with 1400 subsampled image bags are used for evaluation.
The key challenge of this dataset is the sparse views with large viewpoint and illumination changes. 
The ETH3D dataset contains 25 indoor and outdoor scenes with sparsely captured high-resolution images and accurately calibrated poses by Lidar as ground truth.
The proposed Texture-Poor SfM dataset contains low-textured object-centric scenes with 1020 subsampled image bags in total.
On all datasets, images are considered unordered for all methods.

\subsection{Implementation Details}
Our detector-free SfM framework is implemented with multiple detector-free matchers, including LoFTR~\cite{Sun2021LoFTRDL}, MatchFormer~\cite{wang2022matchformer} and AspanTransformer~\cite{Chen2022ASpanFormerDI}, to demonstrate the compatibility of our pipeline.
In the coarse SfM phase, we use their coarse-level matches~($r=8$) as quantized matches for SfM~\cite{schonbergerStructurefromMotionRevisited2016}.
Then, the refinement is performed twice.
A maximum of $16$ views are used for multi-view refinement matching, where longer tracks will be divided into segments and processed separately.
The local patch size for feature extraction $p=15$ and the region size for reference location search $w=7$.
The backbone from S2DNet~\cite{Germain2020S2DNetLA} is used as the CNN feature extractor, and the number of attention groups $n=2$.
The linear attention~\cite{katharopoulos_et_al_2020} is used in all attention layers for efficiency.
In geometry refinement, the BA and topology adjustment are alternated five times, and the outlier filter threshold $\epsilon=3px$.
The running time reported in the experiments is measured using four NVIDIA-V100 GPUs for parallelized matching and 16 CPU cores for BA.

\subsection{Multi-View Camera Pose Estimation}
Camera pose estimation is a central goal of SfM.
This section evaluates the recovered multi-view poses.

\paragraph{Evaluation Protocols.}
On all datasets, matches are built exhaustively between all tentative image pairs, and the same image resizing strategy is used for all methods.
For all the baselines, the default hyperparameters in their original implementations are used.
The AUC of pose error at different thresholds is used as the metric to evaluate the accuracy of estimated multi-view poses, following the IMC benchmark~\cite{jin2021image} and PixSfM~\cite{Lindenberger2021PixelPerfectSW}.
More details are provided in the supplementary material.
\paragraph{Results.}
As shown in Tab.~\ref{tab:exp camerapose}, our detector-free SfM framework outperforms existing baselines over all datasets.
On the ETH3D dataset with high-resolution images, our framework with LoFTR achieves the highest multi-view pose accuracy.
Even when detector-based methods are further refined with PixSfM for multi-view consistency, our framework still surpasses them by a large margin.
On the IMC dataset with large viewpoint and illumination changes, the detector-based baseline SP+SG+PixSfM achieves remarkable performance, while our detector-free framework consistently performs better on all metrics.
The results demonstrate the robustness and effectiveness of our framework on large-scale outdoor scenes with internet images.
Due to the severe low-textured scenario and viewpoint changes in the Texture-Poor SfM dataset, detector-based methods struggle with poor keypoint detection, as shown in Fig.~\ref{fig:reconresults}.
Thanks to the detector-free design, our framework achieves significantly higher accuracy.

Compared with LoFTR+PixSfM, the detector-free baseline that uses the same LoFTR coarse matches as ours, our framework is more accurate on all datasets and metrics, especially on the AUC@1$\degree$ metric with a strict error threshold, which demonstrates the effectiveness of our iterative refinement pipeline with the multi-view matching module.

\subsection{3D Triangulation}
\begin{table}[t]
    \centering
    \resizebox{1.0\columnwidth}{!}{
    \setlength\tabcolsep{6pt} %
    \begin{tabular}{cccccccc} 
    \toprule
    &\multirow{2}{*}{Method}         & \multicolumn{3}{c}{Accuracy~($\%$)}             & \multicolumn{3}{c}{Completeness~($\%$)} \\ 
    \cmidrule(lr){3-5}
    \cmidrule(lr){6-8}
                            & & 1cm       & 2cm       & 5cm       & 1cm       & 2cm       &  5cm \\ 
    \midrule
    \multirow{4}{*}{\begin{tabular}[c]{@{}c@{}}Detector-\\ Based\end{tabular}} & SIFT + NN + PixSfM & 76.18 & 85.60 & 93.16 & 0.17 & 0.71 & 3.29\\
    &D2Net + NN + PixSfM & 74.75 & 83.81 & 91.98 & 0.83 & 2.69 & 8.95\\
    &R2D2 + NN + PixSfM & 74.12 & 84.49 & 91.98 & 0.43 & 1.58 & 6.71\\
    &SP + SG + PixSfM & 79.01 & 87.04 & 93.80 & 0.75 & 2.77 & 11.28 \\
    \midrule
    \multirow{5}{*}{\begin{tabular}[c]{@{}c@{}}Detector-\\ Free\end{tabular}} & LoFTR + PatchFlow & 66.73 & 78.73 & 89.93 & 3.48 & 11.34 & 30.96 \\ 
    & LoFTR + PixSfM & 74.42 & 84.08 & 92.63 & 2.91 & 9.39 & 27.31 \\
    &Ours~(LoFTR) &  \textbf{80.38} & \textbf{89.01} & \textbf{95.83} & 3.73 & 11.07 & 29.54 \\ 
    &Ours~(AspanTrans.) & 77.63 & 87.40 & 95.02 & \textbf{3.97} & \textbf{12.18} & \textbf{32.42}\\ 
    &Ours~(MatchFormer) & 79.86 & 88.51 & 95.48 & 3.76 & 11.06 & 29.05\\
    \bottomrule
    \end{tabular}
    }

    \vspace{0.15cm}
    \caption{\textbf{Results of 3D Triangulation.}
    Our method is compared with the baselines on the ETH3D~\cite{Schps2017AMS} dataset using accuracy and completeness metrics with different thresholds.
    }
    \label{tab:exptriangulation}
    \vspace{-0.35 cm}
    \end{table}
With known camera poses and intrinsics, triangulating accurate scene point clouds based on image correspondences is another important task in SfM.
This section evaluates the accuracy and completeness of triangulated point clouds.

\paragraph{Evaluation Protocols.}
The training set of ETH3D is used for evaluation, which is composed of 13 indoor and outdoor scenes with millimeter-accurate scanned dense point clouds as ground truth.
We follow the protocol used in~\cite{Dusmanu2020MultiViewOO,Lindenberger2021PixelPerfectSW}, which triangulates the scene point clouds with fixed camera poses and intrinsics.
Then, we use the ETH3D benchmark~\cite{Schps2017AMS} to evaluate the triangulated point clouds in terms of accuracy and completeness.
The metrics are reported with different distance thresholds including $(1cm, 2cm, 5cm)$, which are averaged across all scenes.
The results of SIFT, D2Net, and R2D2 descriptors are from the PixSfM~\cite{Lindenberger2021PixelPerfectSW} paper, while the results of other baselines are obtained by running their open-source code.

\paragraph{Results.}
The results are presented in Tab.~\ref{tab:exptriangulation}.
Despite the trade-off between accuracy and completeness, our detector-free SfM framework achieves better performances on both metrics.
Compared to the state-of-the-art detector-based baseline SP+SG+PixSfM, our framework with LoFTR coarse matches achieves higher accuracy with $\sim3\times$ reconstruction completeness, thanks to the iterative refinement module.
Our framework with AspanTransformer coarse matches achieves higher completeness while sacrificing a little accuracy compared to using the LoFTR matches.
Compared with the detector-free LoFTR+PixSfM, our method using the same input matches achieves higher performances both in terms of accuracy and completeness, which further demonstrates the effectiveness of our refinement module.

\begin{figure}[tp]
    \centering
    \includegraphics[width=0.95\linewidth]{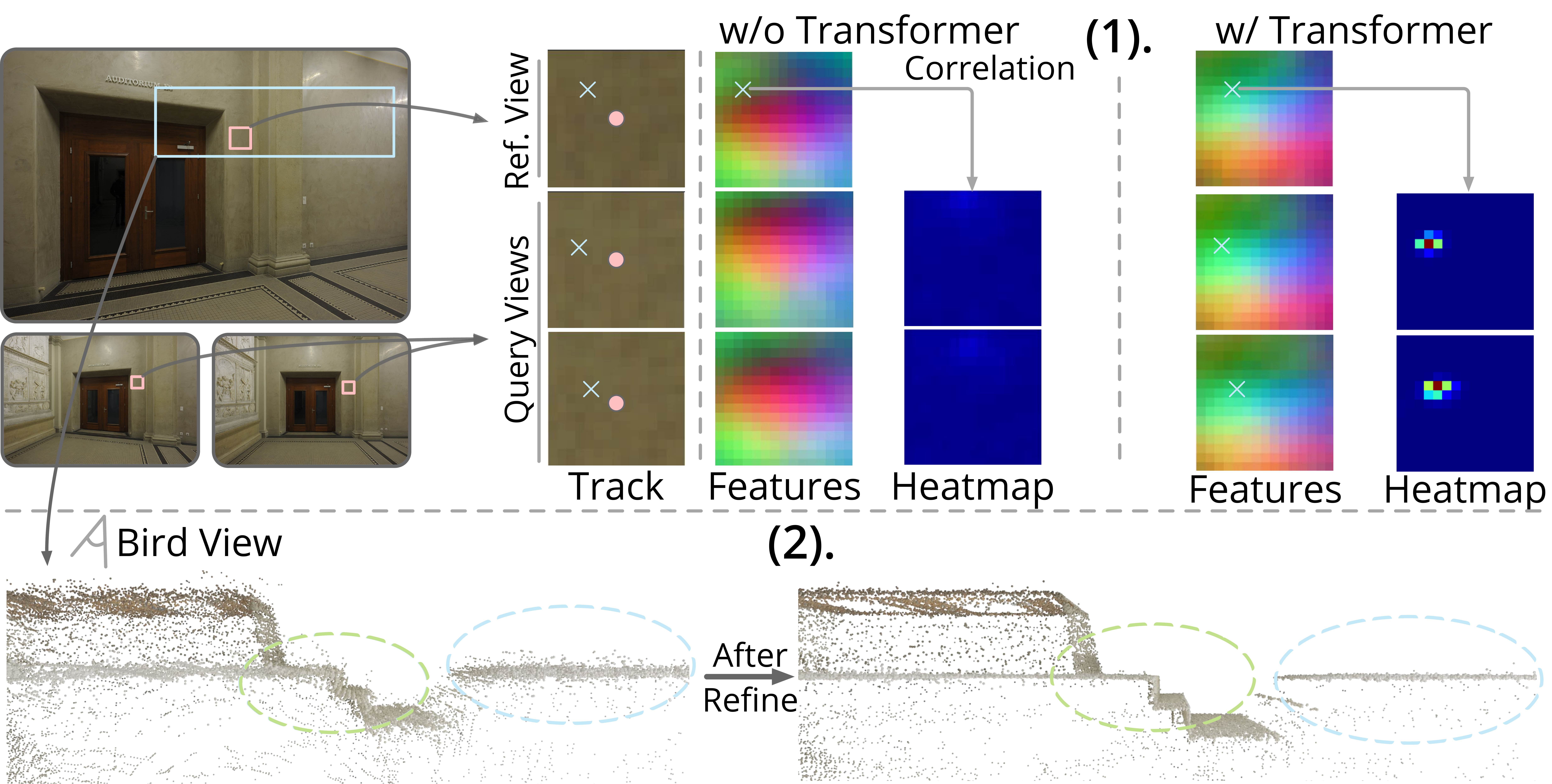}
    \caption{\textbf{Effects of Transformer and Refinement.}
    \textbf{1.} For a feature track in the texture-poor region, its feature patches~(visualized by PCA) become more discriminative after the multi-view transformer.
    \protect\scalerel*{\includegraphics{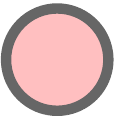}}{B} and \protect\scalerel*{\includegraphics{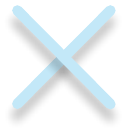}}{B} represent coarse and refined keypoint locations, respectively.
    \textbf{2.} The point cloud after refinement becomes more accurate.
    }
    \vspace{-0.35 cm}
    \label{fig:ablation}
\end{figure}

\subsection{Ablation Studies}
\label{sec:abl}

We conduct several experiments to validate the efficacy of our design choices on the ETH3D dataset with triangulation metrics.
More ablation studies with pose metrics are in the supplementary material.

\paragraph{Coarse Match Quantization.}
Tab.~\ref{tab:ablation}~(1) shows the impact of the match quantization rounding ratio $r$.
Our framework achieves satisfying accuracy and completeness directly using the coarse-level matches output by LoFTR ($r=8$).
Using a smaller quantization ratio yields better matching accuracy but significantly more 2D and 3D points, thus decreasing running efficiency.

\paragraph{Number of Refinement Iterations.}
Tab.~\ref{tab:ablation}~(2) reports the results after each refinement iteration.
Without refinement, the coarse SfM point cloud is inaccurate due to the match quantization.
After the first iteration, the accuracy improves significantly, especially on the \textit{1cm} distance threshold.
Increasing the number of iterations can improve accuracy, with a slight decrease in completeness due to the track merge.
Refining more than twice brings little accuracy improvement while spending more time.
Therefore, we only perform refinement twice for both efficiency and accuracy. 

\paragraph{Maximum Number of Views in Multi-view Matching.}
Tab.~\ref{tab:ablation}~(3) shows the effect of the number of views used for multi-view matching in a single iteration of refinement.
It is shown that using more views for multi-view matching consistently improves both accuracy and completeness without significantly affecting running time.

\begin{table}[t]
    \centering
    \resizebox{1.0\columnwidth}{!}{
    \setlength\tabcolsep{6pt} %
    \begin{tabular}{ccccccc}
    \toprule
    \multirow{2}{*}{}  & \multirow{2}{*}{}         & \multicolumn{2}{c}{Accu.~($\%$)}             & \multicolumn{2}{c}{Complete.~($\%$)} & \multirow{2}{*}{Time~(s)}\\ 
    \cmidrule(lr){3-4}
    \cmidrule(lr){5-6}
        &                    & 1cm       & 2cm       & 1cm       & 2cm  \\ 
    \midrule
    \multirow{3}{*}{(1) Quantization ratio} & $r=8$ & 80.38 & 89.01 & 3.73 & 11.07 & 557 \\
    & $r=4$ &  \textbf{81.58} & \textbf{89.82} & 4.41 & 12.27 & 718\\ 
    & $r=2$ &  81.18 & 89.78 & \textbf{5.41} & \textbf{14.15} & 791\\
    \midrule
    \multirow{4}{*}{(2) Number of iterations} & No refine. & 42.13 & 59.92 & 2.21 & 8.45 & 296\\
    & 1 iter & 77.62 & 87.04 & \textbf{3.83} & \textbf{11.44} & 430 \\
    & 2 iter & 80.38 & 89.01 & 3.73 & 11.07 & 557\\
    & 3 iter & \textbf{81.26} & \textbf{89.59} & 3.57 & 10.64 & 678\\
    \midrule
    \multirow{4}{*}{\begin{tabular}[c]{@{}r@{}}(3) Number of views in\\ multi-view matching\end{tabular}} & 2 views & 69.77 & 81.69 & 2.02 & 7.10 & 438 \\
    & 4 views & 72.30 & 83.42 & 2.48 & 8.35 & 435 \\
    & 8 views & 74.68 & 85.02 & 3.09 & 9.84 & 431 \\
    & 16 views & \textbf{77.62} & \textbf{87.04} & \textbf{3.83} & \textbf{11.44} & 430 \\
    \midrule
    \multirow{4}{*}{(4) Refinement designs}& Full model & \textbf{80.38} & \textbf{89.01} & 3.73 & 11.07 & 557\\
    & w/o transformer & 71.85 & 82.66 & 2.72 & 8.79 & 541\\
    & w/o ref. location search & 76.66 & 86.79 & \textbf{4.16} & \textbf{12.56} & 554\\
    & w/o topology adjustment & 75.58 & 85.47 & 4.07 & 12.17 & 552 \\
    \bottomrule
    \end{tabular}
    }

    \vspace{0.15cm}
    \caption{\textbf{Ablation Studies.}
    On the ETH3D dataset, we quantitatively evaluate the impact of the quantization ratio, the number of iterations of refinement, the number of views used for multi-view matching, and other designs in refinement. The reported triangulation accuracy and completeness are averaged across all scenes, while the running time is evaluated on a single scene \emph{Kicker}.
    }
    \label{tab:ablation}
    \vspace{-0.35 cm}
    \end{table}

\paragraph{Refinement Designs.}
Tab.~\ref{tab:ablation}~(4) shows the benefits of the feature transformer and reference location search in multi-view matching and the track topology adjustment in the geometry refinement.
Compared with multi-view matching that directly uses backbone CNN features for matching, using multi-view transformed features can significantly improve accuracy and completeness.
The result demonstrates the effectiveness of the proposed transformer module, which considers feature relations among multiple views and helps disambiguate features for more accurate matching, as visualized in Fig.~\ref{fig:ablation}~(1).
Reference location search in the reference view brings a $3.7\%$ improvement on the \textit{1cm} metric.
Without the track topology adjustment in geometry refinement, the point clouds' accuracy drops by $4.6\%$ on the strict threshold~(\textit{1cm}), which demonstrates the benefits of topology adjustment on accuracy.

\subsection{Efficiency on Large-Scale Scenes}
\label{subsec scalability}
\begin{figure}[tp]
    \centering
    \includegraphics[width=1.0\linewidth]{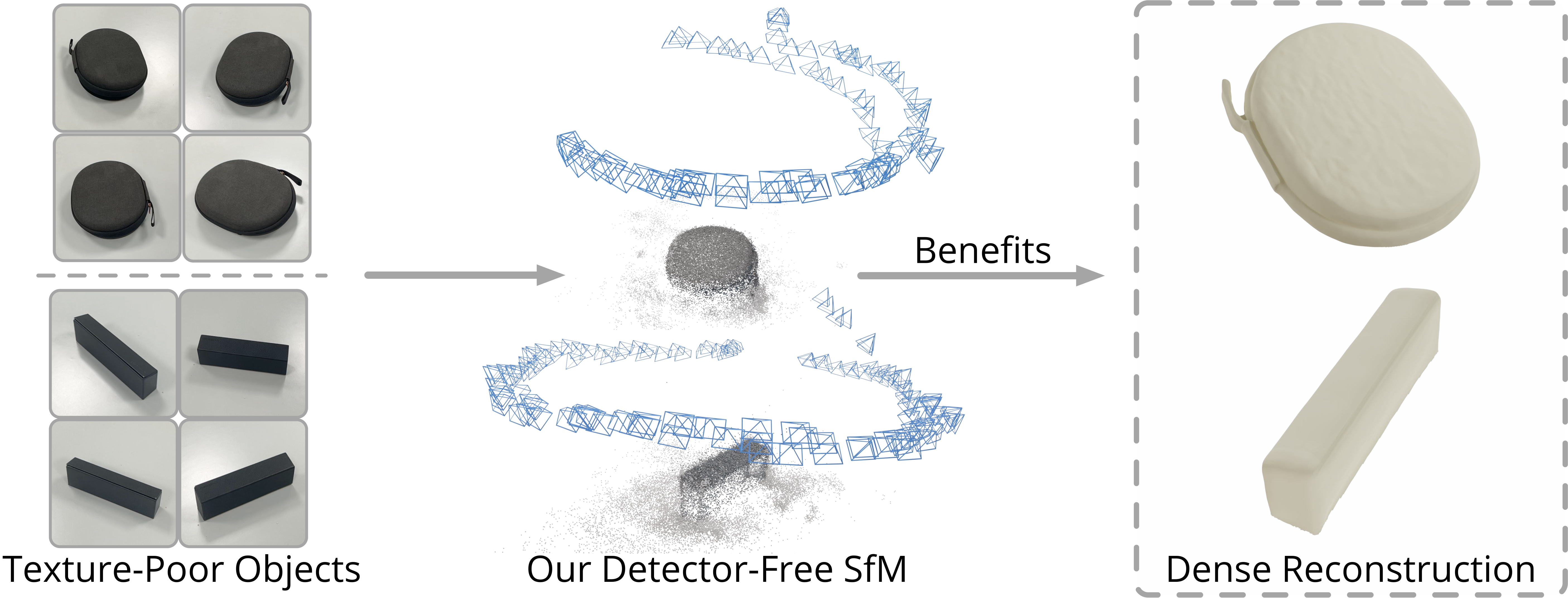}
    \caption{\textbf{Applications.} The recovered poses on texture-poor scenes by our detector-free SfM framework benefit substream tasks, e.g., dense reconstruction using neural implicit fields~\cite{wang2021neus}.
    }
    \vspace{-0.35 cm}
    \label{fig:moreresults}
\end{figure}
We conduct experiments on the Aachen v1.1 dataset~\cite{Sattler2018Benchmarking6O,Zhang2020ReferencePG,Sattler2012ImageRF} to demonstrate the efficiency of our framework in handling large-scale scenes. 
The time and memory costs for refinement are shown in Tab.~\ref{tab:scaleup}. 
We compare our method with PixSfM that uses the same LoFTR~\cite{Sun2021LoFTRDL} coarse matches and the same number of CPU cores as ours, where its cost map approximation is used to reduce the memory footprint and improve efficiency.
Our pipeline achieves competitive efficiency even on the scene with 2000 images and $3.2$ million 3D points.
Moreover, since we perform multi-view matching first and then refine geometry, we do not need to store the feature patch or cost map of each 2D point in memory for BA as PixSfM does.
Therefore, the geometric BA in our pipeline can be very efficient with a small memory footprint on large-scale scenes, which significantly outperforms PixSfM in memory efficiency.

On the scene with 2000 images, the detector-free matching~\cite{Sun2021LoFTRDL} and coarse SfM takes 4.2 hours in total, due to a large number of semi-dense matches and 3D points.
Thus, the overall speed of our framework is slower than detector-based systems that are based on sparse features.
More running time comparisons on large-scale scenes in the 1DSfM~\cite{Wilson2014RobustGT} dataset are shown in the supplementary material.

\begin{table}[t]
    \centering
    \resizebox{1.0\columnwidth}{!}{
    \setlength\tabcolsep{6pt} %
    \begin{tabular}{ccccc} 
    \toprule
    \multicolumn{2}{c}{} & 500 Images       & 1000 Images       & 2000 Images\\ 
    \midrule
    \multicolumn{2}{c}{Number of 3D Points} & 553k & 1525k & 3235k \\
    \midrule
    \multicolumn{2}{c}{Ours Refinement Time~(s)} & 312 & 969 & 2319\\
    \midrule
    \multirow{3}{*}{\begin{tabular}[c]{@{}c@{}}BA\\ Memory~(GB)\end{tabular} } & PixSfM~(Feature Map) & 161.7 & 393.8 & 904.5 \\
    & PixSfM~(Cost Map)& 3.79 & 9.23 & 21.2\\
    & Ours & \textbf{0.37} & \textbf{1.21} & \textbf{2.63}\\ %
    \bottomrule
    \end{tabular}
    }

    \vspace{0.15cm}
    \caption{\textbf{Efficiency on Large-Scale Scenes.}
    Our method is compared with PixSfM. Both of them use LoFTR coarse matches as input and share the same coarse SfM initialization. Only refinement time and peak memory footprint during BA are reported.
    }
    \label{tab:scaleup}
    \vspace{-0.55 cm}
    \end{table}
\section{Conclusions}
We propose a new detector-free SfM framework to recover camera poses and point clouds from unordered images.
In contrast to traditional SfM systems that depend on keypoint detection at the beginning, our framework leverages the recent success of detector-free matchers to avoid early determination of keypoints which may break down the whole SfM system if the detected keypoints are not repeatable, which often occur in challenging texture-poor scenes. 
Extensive experiments demonstrate that our framework outperforms detector-based SfM baselines across all datasets and metrics. 
We believe the proposed SfM framework opens up the possibility to reconstruct texture-poor scenes from unordered images as shown in Fig.~\ref{fig:overview} and Fig.~\ref{fig:reconresults} and benefits downstream tasks such as dense reconstruction and view synthesis as shown in Fig.~\ref{fig:moreresults}, as it can recover accurate poses and relatively dense point clouds.

\paragraph{Limitations and future works.}
The main limitation of our framework is efficiency.
Due to the significant number of matches produced by detector-free matches, the overall mapping phase will be inevitably slower than the previous detector-based pipelines, especially on large-scale scenes.

As future work, our framework can be extended with more advanced parallelized BA methods~\cite{Huang2021DeepLMLN,2021megba} for better efficiency and integration with multi-modality data such as depth maps and IMUs if available in real applications.

{
   \small
   \bibliographystyle{ieee_fullname}
   \bibliography{egbib}

\begin{thebibliography}{10}\itemsep=-1pt

\bibitem{arkit}
{ARK}it.
\newblock \url{https://developer.apple.com/augmented-reality/}.

\bibitem{Metashape}
Metashape.
\newblock \url{https://www.agisoft.com/}.

\bibitem{realtycapture}
Reality capture.
\newblock \url{https://www.capturingreality.com/}.

\bibitem{Agarwal2009BuildingRI}
Sameer Agarwal, Yasutaka Furukawa, Noah Snavely, Ian Simon, Brian Curless,
  Steven~M. Seitz, and Richard Szeliski.
\newblock Building rome in a day.
\newblock {\em ICCV}, 2009.

\bibitem{Agarwal2010BundleAI}
Sameer Agarwal, Noah Snavely, Steven~M. Seitz, and Richard Szeliski.
\newblock Bundle adjustment in the large.
\newblock In {\em ECCV}, 2010.

\bibitem{Arandjelovi2015NetVLADCA}
Relja Arandjelovi{\'c}, Petr Gron{\'a}t, Akihiko Torii, Tom{\'a}s Pajdla, and
  Josef Sivic.
\newblock Netvlad: Cnn architecture for weakly supervised place recognition.
\newblock {\em TPAMI}, 2015.

\bibitem{Beardsley19963DMA}
Paul~A. Beardsley, Philip H.~S. Torr, and Andrew Zisserman.
\newblock 3d model acquisition from extended image sequences.
\newblock In {\em ECCV}, 1996.

\bibitem{Byrd2010ConjugateGB}
Martin Byr{\"o}d and Kalle {\AA}str{\"o}m.
\newblock Conjugate gradient bundle adjustment.
\newblock In {\em ECCV}, 2010.

\bibitem{Chen2021LearningTM}
Hongkai Chen, Zixin Luo, Jiahui Zhang, Lei Zhou, Xuyang Bai, Zeyu Hu, Chiew-Lan
  Tai, and Long Quan.
\newblock Learning to match features with seeded graph matching network.
\newblock {\em ICCV}, 2021.

\bibitem{Chen2022ASpanFormerDI}
Hongkai Chen, Zixin Luo, Lei Zhou, Yurun Tian, Mingmin Zhen, Tian Fang, David
  N.~R. McKinnon, Yanghai Tsin, and Long Quan.
\newblock Aspanformer: Detector-free image matching with adaptive span
  transformer.
\newblock In {\em ECCV}, 2022.

\bibitem{imagematchingchallenge2023}
Ashley Chow, Eduard Trulls, HCL-Jevster, Kwang~Moo Yi, lcmrll, old ufo, Sohier
  Dane, tanjigou, WastedCode, and Weiwei Sun.
\newblock Image matching challenge 2023, 2023.

\bibitem{Crandall2011DiscretecontinuousOF}
David~J. Crandall, Andrew Owens, Noah Snavely, and Daniel~P. Huttenlocher.
\newblock Discrete-continuous optimization for large-scale structure from
  motion.
\newblock {\em CVPR}, 2011.

\bibitem{Cui2015GlobalSB}
Zhaopeng Cui and Ping Tan.
\newblock Global structure-from-motion by similarity averaging.
\newblock {\em ICCV}, 2015.

\bibitem{DeTone2017SuperPointSI}
Daniel DeTone, Tomasz Malisiewicz, and Andrew Rabinovich.
\newblock Superpoint: Self-supervised interest point detection and description.
\newblock {\em CVPRW}, 2018.

\bibitem{Dusmanu2019CVPR}
Mihai Dusmanu, Ignacio Rocco, Tomas Pajdla, Marc Pollefeys, Josef Sivic,
  Akihiko Torii, and Torsten Sattler.
\newblock {D2-Net: A Trainable CNN for Joint Detection and Description of Local
  Features}.
\newblock In {\em CVPR}, 2019.

\bibitem{Dusmanu2020MultiViewOO}
Mihai Dusmanu, Johannes~L. Sch\"onberger, and Marc Pollefeys.
\newblock {Multi-View Optimization of Local Feature Geometry}.
\newblock In {\em ECCV}, 2020.

\bibitem{Ferrera2019AQUALOCAU}
Maxime Ferrera, Vincent Creuze, Julien Moras, and Pauline Trouv{\'e}-Peloux.
\newblock Aqualoc: An underwater dataset for visual–inertial–pressure
  localization.
\newblock {\em The International Journal of Robotics Research}, 2019.

\bibitem{Fitzgibbon1998AutomaticCR}
Andrew~William Fitzgibbon and Andrew Zisserman.
\newblock Automatic camera recovery for closed or open image sequences.
\newblock In {\em ECCV}, 1998.

\bibitem{Germain2020S2DNetLA}
Hugo Germain, Guillaume Bourmaud, and Vincent Lepetit.
\newblock S2dnet: Learning image features for accurate sparse-to-dense
  matching.
\newblock In {\em ECCV}, 2020.

\bibitem{Gu2021DRODR}
Xiaodong Gu, Weihao Yuan, Zuozhuo Dai, Siyu Zhu, Chengzhou Tang, and Ping Tan.
\newblock Dro: Deep recurrent optimizer for structure-from-motion.
\newblock {\em ArXiv:2103.13201}, 2021.

\bibitem{hampel1986robust}
Frank~R Hampel, Elvezio~M Ronchetti, Peter~J Rousseeuw, and Werner~A Stahel.
\newblock {\em Robust statistics: the approach based on influence functions}.
\newblock Wiley, 1986.

\bibitem{he2022oneposeplusplus}
Xingyi He, Jiaming Sun, Yuang Wang, Di Huang, Hujun Bao, and Xiaowei Zhou.
\newblock Onepose++: Keypoint-free one-shot object pose estimation without
  {CAD} models.
\newblock In {\em NeurIPS}, 2022.

\bibitem{Huang2021DeepLMLN}
Jingwei Huang, Shan Huang, and Mingwei Sun.
\newblock Deeplm: Large-scale nonlinear least squares on deep learning
  frameworks using stochastic domain decomposition.
\newblock {\em CVPR}, 2021.

\bibitem{Jeong2021SelfCalibratingNR}
Yoonwoo Jeong, Seokjun Ahn, Christopher~Bongsoo Choy, Anima Anandkumar, Minsu
  Cho, and Jaesik Park.
\newblock Self-calibrating neural radiance fields.
\newblock {\em ICCV}, 2021.

\bibitem{jin2021image}
Yuhe Jin, Dmytro Mishkin, Anastasiia Mishchuk, Jiri Matas, Pascal Fua,
  Kwang~Moo Yi, and Eduard Trulls.
\newblock Image matching across wide baselines: From paper to practice.
\newblock {\em IJCV}, 2021.

\bibitem{katharopoulos_et_al_2020}
Angelos Katharopoulos, Apoorv Vyas, Nikolaos Pappas, and Fran{\c{c}}ois
  Fleuret.
\newblock Transformers are {RNNs}: Fast autoregressive transformers with linear
  attention.
\newblock In {\em ICML}, 2020.

\bibitem{li20dualrc}
Xinghui Li, Kai Han, Shuda Li, and Victor Prisacariu.
\newblock Dual-resolution correspondence networks.
\newblock In {\em NeurIPS}, 2020.

\bibitem{Li2018MegaDepthLS}
Zhengqi Li and Noah Snavely.
\newblock Megadepth: Learning single-view depth prediction from internet
  photos.
\newblock {\em CVPR}, 2018.

\bibitem{Lin2021BARFBN}
Chen-Hsuan Lin, Wei-Chiu Ma, Antonio Torralba, and Simon Lucey.
\newblock Barf: Bundle-adjusting neural radiance fields.
\newblock {\em ICCV}, 2021.

\bibitem{Lindenberger2021PixelPerfectSW}
Philipp Lindenberger, Paul-Edouard Sarlin, Viktor Larsson, and Marc Pollefeys.
\newblock Pixel-perfect structure-from-motion with featuremetric refinement.
\newblock {\em ICCV}, 2021.

\bibitem{LoweDavid2004DistinctiveIF}
G LoweDavid.
\newblock Distinctive image features from scale-invariant keypoints.
\newblock {\em IJCV}, 2004.

\bibitem{Meng2021GNeRFGN}
Quan Meng, Anpei Chen, Haimin Luo, Minye Wu, Hao Su, Lan Xu, Xuming He, and
  Jingyi Yu.
\newblock Gnerf: Gan-based neural radiance field without posed camera.
\newblock {\em ICCV}, 2021.

\bibitem{Mohr1993Relative3R}
Roger Mohr, Long Quan, and Francoise Veillon.
\newblock Relative 3d reconstruction using multiple uncalibrated images.
\newblock {\em The International Journal of Robotics Research}, 1993.

\bibitem{Parameshwara2022DiffPoseNetDD}
Chethan Parameshwara, Gokul Hari, Cornelia Fermuller, Nitin~J. Sanket, and
  Yiannis Aloimonos.
\newblock Diffposenet: Direct differentiable camera pose estimation.
\newblock {\em CVPR}, 2022.

\bibitem{Pollefeys2004VisualMW}
Marc Pollefeys, Luc~Van Gool, Maarten Vergauwen, Frank Verbiest, Kurt Cornelis,
  Jan Tops, and Reinhard Koch.
\newblock Visual modeling with a hand-held camera.
\newblock {\em IJCV}, 2004.

\bibitem{2021megba}
Jie Ren, Wenteng Liang, Ran Yan, Luo Mai, Shiwen Liu, and Xiao Liu.
\newblock Megba: A gpu-based distributed library for large-scale bundle
  adjustment.
\newblock In {\em ECCV}, 2022.

\bibitem{r2d2}
Jerome Revaud, Philippe Weinzaepfel, C{\'{e}}sar~Roberto de Souza, and Martin
  Humenberger.
\newblock {R2D2:} repeatable and reliable detector and descriptor.
\newblock In {\em NeurIPS}, 2019.

\bibitem{Roberts2011StructureFM}
Richard Roberts, Sudipta~N. Sinha, Richard Szeliski, and Drew Steedly.
\newblock Structure from motion for scenes with large duplicate structures.
\newblock {\em CVPR}, 2011.

\bibitem{Rocco2018NeighbourhoodCN}
I. Rocco, M. Cimpoi, R. Arandjelovi\'c, A. Torii, T. Pajdla, and J. Sivic.
\newblock Neighbourhood consensus networks.
\newblock {\em NeurIPS}, 2018.

\bibitem{Rublee2011ORBAE}
Ethan Rublee, Vincent Rabaud, Kurt Konolige, and Gary~R. Bradski.
\newblock Orb: An efficient alternative to sift or surf.
\newblock {\em ICCV}, 2011.

\bibitem{sarlin20superglue}
Paul-Edouard Sarlin, Daniel DeTone, Tomasz Malisiewicz, and Andrew Rabinovich.
\newblock {SuperGlue}: Learning feature matching with graph neural networks.
\newblock In {\em CVPR}, 2020.

\bibitem{Sattler2018Benchmarking6O}
Torsten Sattler, William~P. Maddern, Carl Toft, Akihiko Torii, Lars
  Hammarstrand, Erik Stenborg, Daniel Safari, M. Okutomi, Marc Pollefeys, Josef
  Sivic, Fredrik Kahl, and Tom{\'a}s Pajdla.
\newblock Benchmarking 6dof outdoor visual localization in changing conditions.
\newblock {\em CVPR}, 2018.

\bibitem{Sattler2012ImageRF}
Torsten Sattler, Tobias Weyand, B. Leibe, and Leif~P. Kobbelt.
\newblock Image retrieval for image-based localization revisited.
\newblock In {\em British Machine Vision Conference}, 2012.

\bibitem{schonbergerStructurefromMotionRevisited2016}
Johannes~L. Sch{\"{o}}nberger and Jan{-}Michael Frahm.
\newblock Structure-from-motion revisited.
\newblock In {\em CVPR}, 2016.

\bibitem{Schps2017AMS}
Thomas Sch{\"o}ps, Johannes~L. Sch{\"o}nberger, S. Galliani, Torsten Sattler,
  Konrad Schindler, Marc Pollefeys, and Andreas Geiger.
\newblock A multi-view stereo benchmark with high-resolution images and
  multi-camera videos.
\newblock {\em CVPR}, 2017.

\bibitem{Shen2022SemiDenseFM}
Zehong Shen, Jiaming Sun, Yuang Wang, Xinying He, Hujun Bao, and Xiaowei Zhou.
\newblock Semi-dense feature matching with transformers and its applications in
  multiple-view geometry.
\newblock {\em TPAMI}, 2022.

\bibitem{Snavely2006PhotoTE}
Noah Snavely, Steven~M. Seitz, and Richard Szeliski.
\newblock Photo tourism: exploring photo collections in 3d.
\newblock {\em TOG}, 2006.

\bibitem{Sun2021LoFTRDL}
Jiaming Sun, Zehong Shen, Yuang Wang, Hujun Bao, and Xiaowei Zhou.
\newblock Loftr: Detector-free local feature matching with transformers.
\newblock {\em CVPR}, 2021.

\bibitem{Tang2019BANetDB}
Chengzhou Tang and Ping Tan.
\newblock Ba-net: Dense bundle adjustment networks.
\newblock In {\em ICLR}, 2019.

\bibitem{tang2022quadtree}
Shitao Tang, Jiahui Zhang, Siyu Zhu, and Ping Tan.
\newblock Quadtree attention for vision transformers.
\newblock {\em ICLR}, 2022.

\bibitem{Triggs1999BundleA}
Bill Triggs, Philip~F. McLauchlan, Richard~I. Hartley, and Andrew~William
  Fitzgibbon.
\newblock Bundle adjustment - a modern synthesis.
\newblock In {\em Workshop on Vision Algorithms}, 1999.

\bibitem{Truong2021LearningAD}
Prune Truong, Martin Danelljan, Luc~Van Gool, and Radu Timofte.
\newblock Learning accurate dense correspondences and when to trust them.
\newblock {\em CVPR}, 2021.

\bibitem{tyszkiewicz2020disk}
Micha{\l} Tyszkiewicz, Pascal Fua, and Eduard Trulls.
\newblock Disk: Learning local features with policy gradient.
\newblock {\em NeurIPS}, 2020.

\bibitem{Vaswani2017AttentionIA}
Ashish Vaswani, Noam~M. Shazeer, Niki Parmar, Jakob Uszkoreit, Llion Jones,
  Aidan~N. Gomez, Lukasz Kaiser, and Illia Polosukhin.
\newblock Attention is all you need.
\newblock In {\em NeurIPS}, 2017.

\bibitem{Vijayanarasimhan2017SfMNetLO}
Sudheendra Vijayanarasimhan, Susanna Ricco, Cordelia Schmid, Rahul Sukthankar,
  and Katerina Fragkiadaki.
\newblock Sfm-net: Learning of structure and motion from video.
\newblock {\em ArXiv:1704.07804}, 2017.

\bibitem{Wang2022TCSfMRT}
Lei Wang, Lin-Lin Ge, Shan Luo, Zi~Jun Yan, Zhaopeng Cui, and Jieqing Feng.
\newblock Tc-sfm: Robust track-community-based structure-from-motion.
\newblock {\em ArXiv}, abs/2206.05866, 2022.

\bibitem{wang2021neus}
Peng Wang, Lingjie Liu, Yuan Liu, Christian Theobalt, Taku Komura, and Wenping
  Wang.
\newblock Neus: Learning neural implicit surfaces by volume rendering for
  multi-view reconstruction.
\newblock In {\em NeurIPS}, 2021.

\bibitem{wang2022matchformer}
Qing Wang, Jiaming Zhang, Kailun Yang, Kunyu Peng, and Rainer Stiefelhagen.
\newblock Matchformer: Interleaving attention in transformers for feature
  matching.
\newblock In {\em ACCV}, 2022.

\bibitem{wang2020CAPS}
Qianqian Wang, Xiaowei Zhou, Bharath Hariharan, and Noah Snavely.
\newblock Learning feature descriptors using camera pose supervision.
\newblock In {\em ECCV}, 2020.

\bibitem{Widya2018StructureFM}
Aji~Resindra Widya, Akihiko Torii, and M. Okutomi.
\newblock Structure from motion using dense cnn features with keypoint
  relocalization.
\newblock {\em IPSJ Transactions on Computer Vision and Applications}, 2018.

\bibitem{Wilson2013NetworkPF}
Kyle Wilson and Noah Snavely.
\newblock Network principles for sfm: Disambiguating repeated structures with
  local context.
\newblock {\em ICCV}, 2013.

\bibitem{Wilson2014RobustGT}
Kyle Wilson and Noah Snavely.
\newblock Robust global translations with 1dsfm.
\newblock In {\em ECCV}, 2014.

\bibitem{Wu2013TowardsLI}
Changchang Wu.
\newblock Towards linear-time incremental structure from motion.
\newblock {\em 3DV}, 2013.

\bibitem{Wu2011MulticoreBA}
Changchang Wu, Sameer Agarwal, Brian Curless, and Steven~M. Seitz.
\newblock Multicore bundle adjustment.
\newblock {\em CVPR}, 2011.

\bibitem{Yan2017DistinguishingTI}
Qingan Yan, Long Yang, Ling Zhang, and Chunxia Xiao.
\newblock Distinguishing the indistinguishable: Exploring structural
  ambiguities via geodesic context.
\newblock {\em CVPR}, pages 152--160, 2017.

\bibitem{Zach2008WhatCM}
Christopher Zach, Arnold Irschara, and Horst Bischof.
\newblock What can missing correspondences tell us about 3d structure and
  motion?
\newblock {\em CVPR}, pages 1--8, 2008.

\bibitem{zhang2022relpose}
Jason~Y. Zhang, Deva Ramanan, and Shubham Tulsiani.
\newblock {RelPose}: Predicting probabilistic relative rotation for single
  objects in the wild.
\newblock In {\em ECCV}, 2022.

\bibitem{Zhang2020ReferencePG}
Zichao Zhang, Torsten Sattler, and Davide Scaramuzza.
\newblock Reference pose generation for long-term visual localization via
  learned features and view synthesis.
\newblock {\em IJCV}, 2020.

\bibitem{Zhou2017UnsupervisedLO}
Tinghui Zhou, Matthew~A. Brown, Noah Snavely, and David~G. Lowe.
\newblock Unsupervised learning of depth and ego-motion from video.
\newblock {\em CVPR}, 2017.

\end{thebibliography}
}

\appendix

\section*{Supplementary Material}
\section{Discussion about Coarse SfM Accuracy}

The coarse SfM phase is NOT to guarantee accuracy. Instead, it sacrifices accuracy for better completeness (registration rate).
To this end, we use \emph{quantized} detector-free matches, where the pixel threshold is relatively large in this phase~(i.e., 4 pixels by default and 8 pixels for high-resolution images in ETH3D dataset) in both of the matching RANSAC and mapping.
Therefore, a sufficient number of images can be registered with an acceptable pose error, which serves as the initialization of the refinement phase for higher pose accuracy.
Nevertheless, Tab.~\ref{tab: coarse pose} shows that the coarse SfM alone can achieve competitive accuracy compared with the state-of-the-art detector-based method on the IMC dataset, resulting in consistently superior results after refinement.
\begin{table}[h]
    \centering
    \resizebox{1.0\columnwidth}{!}{
    \setlength\tabcolsep{3pt} %
    \begin{tabular}{cccccc}
    \toprule
    Type & Method  & AUC@3$\degree$       & AUC@5$\degree$       & AUC@10$\degree$ & AUC@20$\degree$ \\ 
    \midrule
    \multirow{2}{*}{Detector-Based (Reference)} 
    &   R2D2 + NN + PixSfM & 31.41 & 41.80 & 54.65 & 64.90 \\
    &   SP + SG + PixSfM & 45.19  & 57.22 & 70.47 & 79.86 \\
    \hline
    \multirow{2}{*}{Detector-Free} & Ours Coarse SfM (Quant. to 8$\times$8) & 39.88 & 52.42 & 67.16 & 78.14 \\
    & Ours full & \textbf{46.55} & \textbf{58.74} & \textbf{72.19} & \textbf{81.62}\\
    \bottomrule
    \end{tabular}
    }
    \vspace{0.15cm}
    \caption{Results of coarse SfM's pose accuracy on IMC 2021 dataset.}
    \label{tab: coarse pose}
    \end{table}

\section{Method Details}

\subsection{Reference View Selection in Feature Track Refinement}
A track $\mathcal{T}_j = \{\mathbf{x}_k \in \mathbb{R}^2 | k=1:N_j\}$ is divided into reference and query views and features of the reference view are correlated on query views to search for multi-view correspondences.
This strategy avoids exhaustively searching correspondences between every pair within a feature track, which is a complex topology~\cite{Dusmanu2020MultiViewOO} and is inefficient for refinement.
Our criteria for selecting the reference view is to minimize the keypoint scale differences between the reference view and query views to improve the matchability.
Concretely, we define the scale $s_k$ of a 2D observation $\mathbf{x}_k$ in a feature track $\mathcal{T}_j$ as $s_k = \nicefrac{d_k}{f_i}$, where $f_i$ is focal length in intrinsic parameter $\*C_i$ and $d_k$ is the depth of $\mathbf{x}_k$ that is obtained by projecting its corresponding 3D point $\*P_j$ with the current estimated pose $\boldsymbol{\xi}_i$.
Then, the view with a medium scale across the track is selected as the reference view, whereas the rest views are query views.

\subsection{Multi-View Feature Transformer}
The backbone from S2DNet~\cite{Germain2020S2DNetLA} is used as the CNN feature extractor.
We interpolate and fuse the output features of adaption layers in S2DNet, which are at original image resolution and $\nicefrac{1}{8}$ resolution respectively, to create a single feature map.

After the feature extraction, flattening, and concatenation, we use the Linear Transformer~\cite{katharopoulos_et_al_2020} to efficiently transform the reference feature $\tilde{\mathbf{F}}^r$ and query feature $\tilde{\mathbf{F}}^q$.
Linear Transformer reduces the computational complexity of the Transformer~\cite{Vaswani2017AttentionIA} from $O(N^2)$ to $O(N)$ by substituting the exponential kernel with an alternative kernel function $\operatorname{sim}(Q, K) = \phi(Q) \cdot \phi(K)^\mathrm{T}, \text{where}~\phi(\cdot) = \operatorname{elu}(\cdot)+1$.
Please refer to the original paper~\cite{Vaswani2017AttentionIA} for more details.

We denote a set of self- and cross-attention layers as an attention block:
\begin{equation*}
    \begin{cases}
        \tilde{\mathbf{F'}}_{(l+1)}^{r} = \operatorname{SelfAtten}(\tilde{\mathbf{F}}_{(l)}^{r}, \tilde{\mathbf{F}}_{(l)}^{r}) \enspace , \\
        \tilde{\mathbf{F'}}_{(l+1)}^{q} = \operatorname{SelfAtten}(\tilde{\mathbf{F}}_{(l)}^{q}, \tilde{\mathbf{F}}_{(l)}^{q}) \enspace, \\
        \tilde{\mathbf{F}}_{l+1}^{r}, \tilde{\mathbf{F}}_{l+1}^{q} = \operatorname{CrossAtten}(\tilde{\mathbf{F'}}_{(l+1)}^{r}, \tilde{\mathbf{F'}}_{(l+1)}^{q}) \enspace.
    \end{cases}
\end{equation*}
The indices of intermediate features are indicated by ${\cdot}_{(l)}$. $\tilde{\mathbf{F'}}$ represents an intermediate feature processed by a self-attention layer.
Our attention module sequentially performs the attention block $n=2$ times to transform the reference and query features.

The transformed features are reshaped into feature patches $\{\hat{\mathbf{F}}_k \in \mathbb{R}^{p \times p \times c}\}$ for multi-view feature correlation.

\subsection{Geometry Refinement}
With refined feature tracks, we perform bundle adjustment~(BA) to optimize the scene geometry by reprojection error.
The Cauchy function is used as the robust loss function $\rho(\cdot)$.
For efficiency, we form the reduced camera system by the Schur Complement and then solve it by dense or sparse decomposition for small- or medium-scale scenes~(number of images smaller than $500$), respectively.
On large-scale scenes, the reduced camera system is solved by Preconditioned Conjugate Gradients algorithm~(PCG)~\cite{Agarwal2010BundleAI,Byrd2010ConjugateGB}.
Moreover, to reduce the drift during BA, we select the farthest two views in the coarse model and fix the pose of one image and one translation DoF of the other image during BA, following~\cite{schonbergerStructurefromMotionRevisited2016}.

\subsection{Camera Parameter Estimation Details}
Like COLMAP, our method does not require known intrinsic parameters, which can be inferred from image information (EXIF if available, otherwise, using max image edge size as initialization) and refined during BA, both in coarse SfM and refinement phase.
All methods are not provided with intrinsics when evaluated on the IMC and ETH3D datasets.
In the image registration phase, image poses are solved by the PnP algorithm first, followed by no-linear optimization.
Then the poses will be optimized simultaneously with the point cloud in BA.

\section{Training of Multi-View Feature Transformer}
\subsection{Ground Truth Generation}
Our multi-view feature transformation module is trained on the MegaDepth~\cite{Li2018MegaDepthLS}, which is a large-scale outdoor dataset with 196 different scenes.
To construct ground truth feature tracks for training, we first sample image bags for each scene and then project the grid-level points of a randomly selected reference view to other query views by depth maps.

Specifically, we sample $2000$ image bags for each scene with a maximum of six images in each bag. 
The co-visibility extracted from the provided scene SfM model is used to sample image bags. 
We define the co-visibility ratio $v$ of a sampled image bag as:
\begin{equation*}
    v = \frac{|\{\*P\}_0 \cap \{\*P\}_1 \cap \cdots \cap \{\*P\}_i |}{\operatorname{min}(|\{\*P\}_0|, |\{\*P\}_1|,\cdots,|\{\*P\}_i|)} \enspace,
\end{equation*}
where $\{\*P\}_i$ is the set of 3D points observed by image $\*I_i$, and $|\cdot|$ is the operator that calculates the number of elements in a set.
The image bags with a co-visibility ratio $0.02<v<0.6$ are kept for training.
Moreover, the low-quality scenes reported by~\cite{tyszkiewicz2020disk,Dusmanu2020MultiViewOO}~(`0000', `0002', `0011', `0020', `0033', `0050', `0103', `0105', `0143', `0176', `0177', `0265', `0366', `0474', `0860', `4541') and scenes that overlap with IMC test set~(`0024', `0021', `0025', `1589', `0019', `0008', `0032', `0063') are removed from training.

After the image bag sampling, to construct ground truth feature tracks, we randomly select a reference image in the bag and project its grid-level points to other query views by the depth map, intrinsic parameters, and poses.
Since the depth maps in MegaDepth are obtained by the MVS algorithm, inaccurate depth values exist.
For accurate ground-truth multi-view matches, projection depth error and cycle projection error with strict thresholds are checked after the projection to filter inaccurate 2D observations in a feature track.
The projection depth error $e_d$ and cycle projection error $e_c$ are defined as follows:

\begin{equation*}
    \begin{cases}
        e_d = \frac{\| \*D_{q}(\*x_{proj}) - d_{proj} \|}{\*D_{q}(\*x_{proj})} \enspace, \\
        e_c = \| \*x_r -  \boldsymbol{\pi}_r \cdot \boldsymbol{\xi}_{r \rightarrow q}^{-1} \cdot \*D_{q}(\*x_{proj}) \cdot \boldsymbol{\pi}^{-1}_q (\*x_{proj}) \| \enspace,
    \end{cases}
\end{equation*}
\vspace{-1em}
\begin{equation*}
    \text{where} \enspace \*x_{proj}= \boldsymbol{\pi}_q \cdot \boldsymbol{\xi}_{r \rightarrow q} \cdot \*D_r(\*x_r) \cdot \boldsymbol{\pi}^{-1}_r(\*x_r) \enspace.
\end{equation*}
$\*x_r$ is a sampled 2D point in reference view, $\*D_{(\cdot)}$ is the depth map of reference or query view, $\boldsymbol{\pi}$ is the projection determined by intrinsic parameters, and $\boldsymbol{\xi}_{r \rightarrow q} = \boldsymbol{\xi}_{q} \cdot \boldsymbol{\xi}_r^{-1}$ is the relative pose between reference view and a query view. $d_{proj}$ is the $z$ value of 3D points in query view corresponding to $\*x_{proj}$.
A point in the query view is kept in the ground-truth feature track when projection depth error $e_d < 0.005$ and cycle projection error $e_c < 1px$.
\subsection{Loss}
The multi-view transformer module is trained by minimizing the average $\ell_2$ loss on keypoint locations between the refined tracks and the ground-truth tracks.
Following~\cite{wang2020CAPS,Sun2021LoFTRDL}, we make our loss uncertainty weighted with a variance term $\sigma^{2}(\*x)$:
\begin{equation*}
    \mathcal{L}=\frac{1}{N} \sum_{j \in n_t} \sum_{k \in n_v} \frac{1}{\sigma^{2}(\*x)}\left\|\*x-\*x_{gt}\right\|_{2} \enspace,
    \label{eq:loss_fine}
\end{equation*}
where $n_t$ is the number of feature tracks, $n_v$ is the number of query views in a track, and $N$ is the total number of refined keypoints.
$\sigma^{2}(\*x)$ is calculated by the trace of the heatmap's covariance matrix, which is detached during training to prevent the network from decreasing the loss by increasing the variance.

\subsection{Training Details}
The images are resized to have the longest edge of $840$.
The feature backbone is initialized by the pretrained weighted from S2DNet, and the attention blocks are randomly initialized.
We use the AdamW optimizer to train the entire network, where the initial learning rate of backbone and attention blocks are $2\times10^{-4}$ and $4\times10^{-4}$, respectively.
The network training takes about $30$ hours with a batch size of 8 on 8 NVIDIA V100 GPUs.
\section{Texture-Poor SfM Dataset}
In the proposed Texture-Poor SfM dataset, low-textured objects are placed on a texture-less plane, and video is captured surrounding each object.
Each video is recorded at $30$ fps for about 30 seconds in $1920\times1440$ resolution with per-frame poses and intrinsic parameters estimated by ARKit~\cite{arkit}.
To stabilize the feature tracking and pose estimation in ARKit, textured markers are elaborately placed on the plane but far from the object.
Then we annotate the 3D foreground region for later filter backgrounds that are discriminative and can reduce the difficulty of the dataset.

After the data capture, we perform a global BA~\cite{schonbergerStructurefromMotionRevisited2016} to further optimize camera poses estimated by ARKit and reduce the potential drift.
We extract features~\cite{DeTone2017SuperPointSI}, match~\cite{sarlin20superglue} them, and then perform triangulation using the currently estimated poses.
Then the global BA is performed to optimize poses.
The placed discriminative markers can also facilitate feature extraction and matching in this phase.
After the pose refinement, we crop out the background with salient features, and images after crop are resized to $840\times840$.
With the refined poses, we project the annotated foreground regions to each image to filter backgrounds with salient features, where only cropped foreground images without markers are used for evaluation.

To impose larger viewpoint changes, we sample $60$ subset image bags for each scene based on co-visibility, similar to the IMC 2021 dataset~\cite{jin2021image}.
Each bag contains either 5, 10, or 20 images.
\section{Experiments}

\begin{table}[t]
    \centering
    \resizebox{1.0\columnwidth}{!}{
    \setlength\tabcolsep{3pt} %
    \begin{tabular}{cccccccc}
    \toprule
    \multirow{2}{*}{Sparse Det. \& Matcher} & \multirow{2}{*}{Refinement}  & \multicolumn{3}{c}{ETH3D Dataset} & \multicolumn{3}{c}{IMC2021 Dataset} \\
    \cmidrule(lr){3-5}
    \cmidrule(lr){6-8}
    & & AUC@3$\degree$       & AUC@5$\degree$       & AUC@10$\degree$ & AUC@3$\degree$       & AUC@5$\degree$       & AUC@10$\degree$\\ 
    \midrule
    \multirow{2}{*}{SIFT + NN} & PixSfM & 26.94 & 39.01 & 42.19 & 25.54 & 34.80 & 46.73 \\
    &   Ours & \textbf{29.28} & \textbf{41.76} & \textbf{45.12} & \textbf{26.77} & \textbf{36.18} & \textbf{48.32}  \\
    \midrule
    \multirow{2}{*}{R2D2 + NN} & PixSfM & 43.58 & 62.09 & 66.89 & 31.41 & 41.80 & 54.65 \\
    &   Ours & \textbf{46.84} & \textbf{64.31} & \textbf{68.75} & \textbf{32.35} & \textbf{42.83} & \textbf{55.65}   \\
    \midrule
    \multirow{2}{*}{SP + SG} & PixSfM & 50.82 & 68.52 & 72.86 & 45.19  & 57.22 & 70.47 \\
    &   Ours & \textbf{52.66} & \textbf{70.15} & \textbf{74.85} & \textbf{45.43} & \textbf{57.75} & \textbf{71.57} \\
    \bottomrule
    \end{tabular}
    }
    \vspace{0.15cm}
    \caption{Comparison of sparse local features accompanied with our refinement and PixSfM on ETH3D dataset and IMC2021 dataset.}
    \label{tab: spwithours}
    \end{table}
\subsection{Datasets}
On the IMC dataset, the validation set is already separated.
We follow their protocol and use all eight test scenes for evaluation, and use validation scenes \emph{Sacre Coeur}, \emph{Saint Peter's Square}, and \emph{Reichstag} for tuning hyperparameters.
Images are resized so that the longest edge dimension is equal to 1200 pixels for all methods.
As for the ETH3D dataset, the images are resized to have a maximum edge dimension of 1600 pixels for all methods.
On the proposed Texture-Poor SfM dataset, we use randomly selected three scenes as validation sets for tuning hyper-parameters and the remaining scenes for evaluation.
Note that due to the image crop in the post-process of the Texture-Poor SfM dataset, the principle points of intrinsic parameters are not in the image center.
To avoid the degeneration of estimating principle points in SfM, all of the methods are provided with known intrinsic parameters, which are kept fixed during SfM.

\subsection{Metric of Multi-View Camera Pose Estimation}
The AUC of pose error at different thresholds is used as the metric to evaluate the accuracy of estimated multi-view poses, following the IMC benchmark~\cite{jin2021image} and PixSfM~\cite{Lindenberger2021PixelPerfectSW}.
This metric converts $N$ multi-view poses to $C_N^2$ pair-wise relative transformation, which is invariant to the difference of coordinate system between reconstructed and ground-truth poses.
The pose error is defined as the maximum angular error in rotation and translation.
    
On the IMC dataset and Texture-Poor SfM dataset, we use pose error at $(3\degree, 5\degree, 10\degree)$ thresholds.
Since the ETH3D dataset has high-resolution images and accurate ground truth calibration, we further report a more strict $1\degree$ threshold to evaluate the capability of highly accurate pose estimation.

\subsection{Sparse Features with Our Refinement.}
Our refinement module can also be used to refine SfM models reconstructed by sparse feature detecting and matching to further bring pose improvement. Pose accuracy is evaluated on the ETH3D dataset and IMC 2021 dataset. Results shown in Tab.~\ref{tab: spwithours} demonstrate that our framework can consistently outperform PixSfM when accompanied by the same sparse detectors and matchers.

\subsection{More Ablation Studies}
In this part, we validate the effectiveness of our refinement pipeline by the multi-view pose metric on multiple datasets.
\begin{table}[t]
    \centering
    \resizebox{1.0\columnwidth}{!}{
    \setlength\tabcolsep{8pt} %
    \begin{tabular}{cccccccc}
    \toprule
    \multirow{2}{*}{} & \multirow{2}{*}{}    & \multicolumn{3}{c}{ETH3D Dataset} & \multicolumn{3}{c}{IMC~(\emph{Mount Rushmore})} \\ 
    \cmidrule(lr){3-5}
    \cmidrule(lr){6-8}
         & & AUC@1$\degree$       & AUC@3$\degree$       & AUC@5$\degree$  & AUC@3$\degree$       & AUC@5$\degree$       & AUC@10$\degree$\\ 
    \midrule
    \multirow{3}{*}{LoFTR~\cite{Sun2021LoFTRDL}} & No Refine & 30.88 & 58.90 & 68.06 & 21.26 & 32.09 & 47.96\\
    & Iter 1 & 57.20 & 74.61 & 78.85 & 29.69 & 41.42 & 56.61\\
    & Iter 2 & \textbf{59.12} & \textbf{75.59} & \textbf{79.53} & \textbf{32.35} & \textbf{43.92} & \textbf{58.91}\\
    \midrule
    \multirow{3}{*}{AspanTrans.~\cite{Chen2022ASpanFormerDI}} & No Refine & 28.41 & 55.87 & 65.40 & 19.04 & 29.02 & 44.26\\
    & Iter 1 &  55.48 & 72.84 & 77.07 & 29.06 & 40.29 & 55.11\\
    & Iter 2 & \textbf{57.23} & \textbf{73.71} & \textbf{77.70} & \textbf{31.77} & \textbf{43.23} & \textbf{57.79}\\
    \midrule
    \multirow{3}{*}{MatchFormer~\cite{wang2022matchformer}} & No Refine & 26.30 & 53.95 & 63.48 & 8.48 & 15.46 & 28.47\\
    & Iter 1 & 54.30& 71.29 & 75.52 & 24.25 & 35.10 & 49.80\\ 
    & Iter 2 & \textbf{56.70}& \textbf{73.00} & \textbf{76.84} & \textbf{29.31} & \textbf{39.66} & \textbf{53.32}\\
    \bottomrule
    \end{tabular}
    }
    \vspace{0.15cm}
    \caption{\textbf{Ablation Study of Refinement Iterations.}
    On the ETH3D dataset and scene \emph{Mount Rushmore} in the IMC dataset, we quantitatively evaluate the impact of the number of refinement iterations. 
    The AUC of pose error at different thresholds is reported.
    }
    \label{tab:ablrefine}
    \end{table}
\begin{table}[t]
    \centering
    \resizebox{0.75\columnwidth}{!}{
    \setlength\tabcolsep{6pt} %
    \begin{tabular}{cccc}
    \toprule
    \multirow{2}{*}{}  & \multicolumn{3}{c}{IMC~(\emph{Mount Rushmore})} \\ 
    \cmidrule(lr){2-4}
         & AUC@3$\degree$       & AUC@5$\degree$       & AUC@10$\degree$\\ 
    \midrule
    Medium Scale & \textbf{32.35} & \textbf{43.92} & \textbf{58.91} \\
    Smallest Scale & 30.63 & 42.11 & 56.99 \\
    Largest Scale & 31.94 & 42.98 & 57.54 \\
    Random Selection & 31.21 & 42.45 & 57.24 \\
    \bottomrule
    \end{tabular}
    }
    \vspace{0.15cm}
    \caption{\textbf{Ablation Study of Reference View Selection.}
    On the scene \emph{Mount Rushmore} in the IMC dataset, we evaluate the impact of the reference view selection strategies.
    The AUC of pose error at different thresholds is reported.
    }
    \label{tab:ablref}
    \end{table}
The results in Tab.~\ref{tab:ablrefine} indicate that our iterative refinement pipeline consistently improves pose accuracy for various detector-free matchers across different datasets.
As shown in Tab~\ref{tab:ablref}, using other reference view selection strategies, including using a view with the smallest or largest scale, and randomly selecting a reference view for each track, will reduce the final pose accuracy.

\subsection{Efficiency on Large-Scale Scenes}
Experiments of the large-scale scenes are conducted on a server using 16 CPU cores~(Intel Xeon Gold 6146) and four NVIDIA V100 GPUs.
The subset images are uniformly sampled from the Aachen v1.1 dataset~\cite{Sattler2018Benchmarking6O,Zhang2020ReferencePG,Sattler2012ImageRF}.
For each image, the top 20 most covisible images determined by image retrieval~\cite{Arandjelovi2015NetVLADCA} are used for matching, where images are resized so that the longest edge equals $1200$.
To refine the large-scale scene with a large number of 3D points caused by semi-dense matchers, we perform refinement only once and use four GPUs for parallelized multi-view matching.
As for geometry refinement, we use multi-core bundle adjustment~\cite{Wu2011MulticoreBA} to leverage multiple CPU cores.

\begin{figure}[t]
    \centering
    \includegraphics[width=1.0\linewidth]{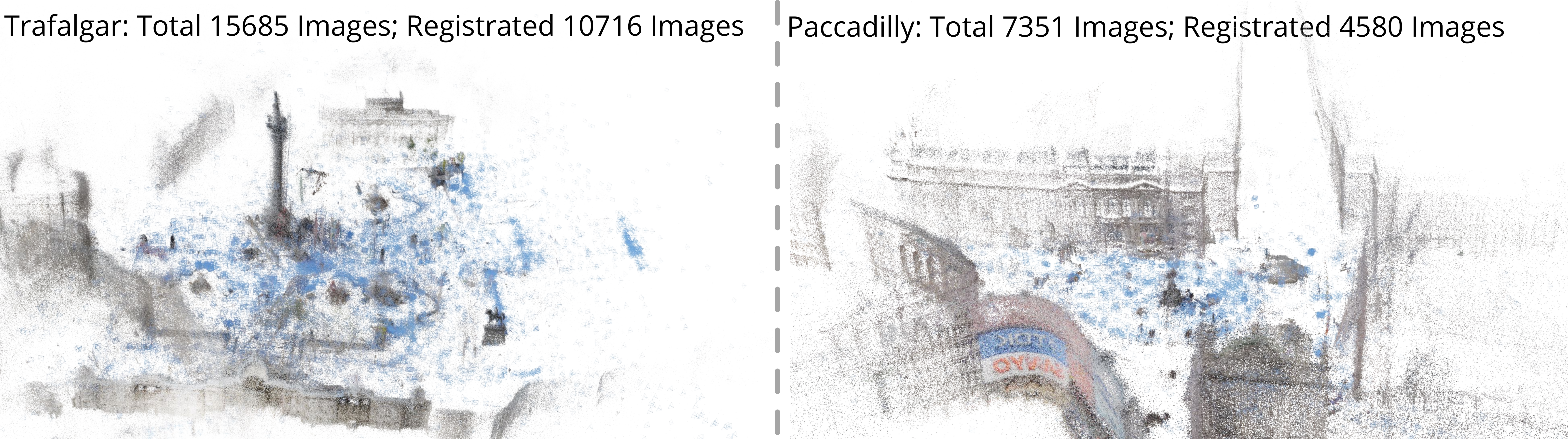}
    \caption{Reconstruction of scenes in the 1DSfM dataset.}
    \label{fig: 1dsfm}
\end{figure}
\paragraph{Comparison with detector-based pipeline.}
We show the overall running time comparison with detector-based pipelines on the four largest scenes in the 1DSfM~\cite{Wilson2014RobustGT} dataset in Tab.~\ref{tab: 1DSfM}.
On large-scale scenes, our framework is slower than detector-based pipeline with sparse features.
On the one hand, detector-free matching is inherently slower than sparse matching. Moreover, due to a significant number of matches produced by detector-free matchers, the incremental mapping phase is also slower.
However, on the scene with images collected from the internet and with large viewpoint and illumination changes, our framework can register significantly more images compared with sparse methods. These results also show that our framework applies to large-scale scenes (with more than $15000$ images). Visualizations of reconstruction are shown in Fig.~\ref{fig: 1dsfm}

\begin{table}[t]
    \centering
    \resizebox{1.0\columnwidth}{!}{
    \setlength\tabcolsep{1pt} %
    \begin{tabular}{ccccccccc} 
    \toprule
    \multirow{2}{*}{Method}         & \multicolumn{2}{c}{Trafalgar (15685 Img)} & \multicolumn{2}{c}{Piccadilly (7351)} & \multicolumn{2}{c}{Vienna Cathe. (6288)} & \multicolumn{2}{c}{Union Squere (5960)} \\ 
    \cmidrule(lr){2-3}
    \cmidrule(lr){4-5}
    \cmidrule(lr){6-7}
    \cmidrule(lr){8-9}
     & Num. Img. & Time & Num. Img. & Time & Num. Img. & Time & Num. Img. & Time\\
    \midrule
    COLMAP & 7001  & \textbf{3.2h} & 2950 & \textbf{1.8h} & 1139 & \textbf{1.1h} & 1026 & \textbf{0.9h} \\
    COLMAP (SP + SG) & 9482 & 6.8h & 3488 & 3.4h & 1764 & 2.7h & 1835 & 2.5h \\
    Ours &  \textbf{10716} & 19.7h  & \textbf{4580} & 11.2h  & \textbf{2436} & 8.8h & \textbf{2026} & 8.2h \\
    \bottomrule
    \end{tabular}
    }
    \vspace{0.15cm}
    \caption{Comparsion with detector-based methods on the 1DSfM dataset.}
    \label{tab: 1DSfM}
    \end{table}
\section{Failure Cases}
\begin{figure}[t]
    \centering
    \includegraphics[width=1.0\linewidth]{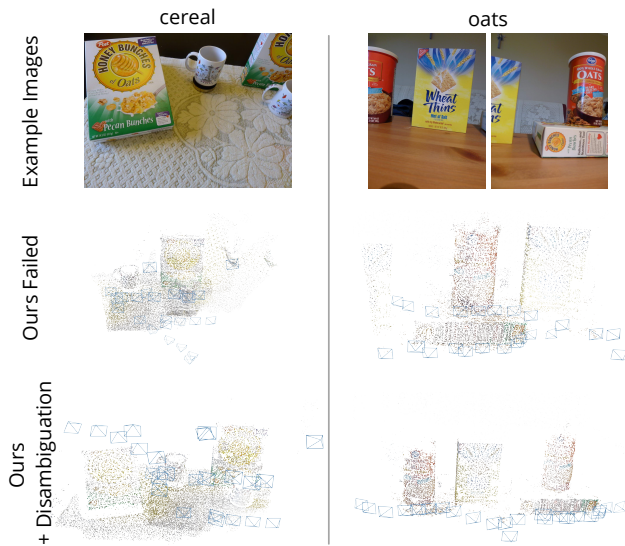}
    \caption{Failure cases of our framework on scenes~\cite{Roberts2011StructureFM} with strong duplicated structures. With the help of disambiguation method~\cite{Cui2015GlobalSB}, our framework can correctly reconstruct ambiguous scenes.}
    \label{fig: failurecases}
\end{figure}

As shown in Fig.~\ref{fig: failurecases}, on the scenes~\cite{Roberts2011StructureFM} with strong duplicated structures, our framework may yield error registrations, which may come with the side effects of detector-free matchers that are capable of matching texture-poor scenes.
Many previous methods~\cite{Zach2008WhatCM,Roberts2011StructureFM,Cui2015GlobalSB,Yan2017DistinguishingTI,Wang2022TCSfMRT} have focused on solving scene disambiguations, which can be further integrated into our framework to alleviate this problem.
With the Missing Correspondence Disambiguation~\cite{Zach2008WhatCM,Cui2015GlobalSB}, our framework can successfully reconstruct the ambiguous scenes, as shown in Fig.~\ref{fig: failurecases}~(row 3).

\section{Real-World Scenes ``Deep Sea'' and ``Moon Surface''}
In this section, we introduce the data collection and running of challenging real-world scenes shown in the main paper's Fig.~\textbf{1} and the demo video.
The \emph{Deep Sea} scene is from sequence 5 of the Aqualoc dataset~\cite{Ferrera2019AQUALOCAU}.
This sequence was chosen because it contains a texture-poor section that presents significant challenges.
The \emph{Moon Surface} sequence is taken from an internet video that has low-texture and repetitive patterns, as well as severe motion blur.

For the scene \emph{Deep Sea}, we use the image retrieval~\cite{Arandjelovi2015NetVLADCA} to select the top 30 most covisible images of each image for matching.
As for the \emph{Moon Surface}, we use sequential matching to match an image with its nearest $20$ frames and run our framework.

\section{Dense Reconstruction}
To demonstrate the application of our framework that can provide accurate poses for dense reconstruction on texture-poor scenes, we run our framework on the scene \emph{Headphone Box} and \emph{Eyeglass Box}.
The sequences are downsampled to 6fps for running our detector-free SfM framework, which provides the recovered poses for neural surface reconstruction method NeuS~\cite{wang2021neus} to reconstruct the scene.
We manually filter the background reconstruction and only keep the object of interest for visualization.

\end{document}